%% file: main.tex
\pdfoutput=1
\PassOptionsToPackage{table}{xcolor}
\documentclass[10pt,a4paper,logo]{googledeepmind}

\usepackage{times}
\usepackage{multirow}
\usepackage{array}
\usepackage{mathrsfs}
\usepackage{mathtools}
\usepackage[numbers,sort&compress]{natbib}
\usepackage{etoolbox}

\AtBeginEnvironment{tabular}{\small}
\AtBeginEnvironment{tabular*}{\small}
\AtBeginEnvironment{tabularx}{\small}

\title{Quantization Degradation in Large Language Models: A Signal--Noise Perspective}

\author[1,2]{Chenxi Zhou}
\author[1,3]{Pengfei Cao\textsuperscript{\dag}}
\author[1]{Jinyu Ye}
\author[1,2]{Bohan Yu}
\author[1]{Haida Yu}
\author[4]{Jiang Li}
\author[1,3]{Jun Zhao}
\author[1,3]{Kang Liu\textsuperscript{\dag}}

\affil[1]{The Key Laboratory of Cognition and Decision Intelligence for Complex Systems, Institute of Automation,\protect\\ Chinese Academy of Sciences, Beijing, China}
\affil[2]{School of Advanced Interdisciplinary Sciences, University of Chinese Academy of Sciences, Beijing, China}
\affil[3]{School of Artificial Intelligence, University of Chinese Academy of Sciences, Beijing, China}
\affil[4]{College of Computer Science, Inner Mongolia University, Hohhot, China}
\correspondingauthor={pengfei.cao@nlpr.ia.ac.cn; kliu@nlpr.ia.ac.cn}

\begin{abstract}
Post-training quantization reduces the deployment cost of large language models, yet how severely a quantized model degrades is not determined by bit-width alone. We systematically study weight-only post-training quantization across bit-widths, quantization methods, model scales and downstream tasks on multiple model families. We observe that such degradation varies substantially across these factors: 4-bit quantization usually preserves performance, 2-bit often causes broad degradation, and at 3-bit, degradation becomes apparent but varies markedly with task type, quantization method and model scale. To explain this variability, we use the signal-to-noise ratio (SNR) to measure how strongly quantization perturbs full-precision representations. We trace degradation back to two linked processes: how quantization errors arise within individual modules, and how they accumulate across layers. First, a source SNR decomposition shows that newly introduced errors depend on three factors: the magnitude of the weight error, the strength of the task-specific signal, and how strongly the quantization error aligns with task-specific activations. Different factors affect these components in distinct ways. Second, a cross-layer propagation analysis shows that these errors can be attenuated, preserved, or amplified as they pass across layers, and that larger models benefit from weaker error amplification. Together, these results establish that quantization degradation is governed by how errors are introduced at the source and how they accumulate across the network.
\end{abstract}

\begin{document}
\maketitle

\section{Introduction}

Large language models (LLMs) are increasingly deployed in settings where memory, latency, and computational budgets are tightly constrained. Model quantization has consequently become a widely adopted approach for reducing deployment costs by representing weights or activations with lower numerical precision while preserving much of the original performance~\cite{gong2025survey}. Among quantization approaches, post-training quantization (PTQ) is particularly attractive because it compresses pretrained LLMs without requiring costly retraining~\cite{frantar2023iclr,lin2024mlsys,dettmers2024spqr}. Weight-only PTQ compresses model weights while keeping activations in high precision during inference. This makes it especially relevant for large-scale deployment, as it directly reduces weight storage and memory bandwidth while avoiding the engineering complexity of activation quantization. At moderate precision levels, such as 4-bit, this strategy often achieves a favourable trade-off between efficiency and performance retention~\cite{jin2024acl}.

Although bit-width is commonly used to describe quantization strength, it is not a reliable predictor of whether a quantized model retains performance or suffers severe degradation as precision is further reduced~\cite{liu2025nips,zhao2025benchmarking,zhou2026taskstratified}. Models compressed to the same bit-width can exhibit markedly different reliability depending on how large the model is, how it is compressed, and what task it performs~\cite{li2024icml,liu2024evaluating,lee2024comprehensive,zhao2025benchmarking}. For example, under 3-bit GPTQ quantization~\cite{frantar2023iclr}, the average zero-shot accuracy across five benchmarks falls from 73.71 to 59.18 for Llama-3.1-8B-Instruct, but only from 78.99 to 77.01 for Llama-3.1-70B-Instruct~\cite{dadgarnia2026gsq}. Such observations indicate that quantization degradation is not determined by bit-width alone. Rather, model scale, quantization method, and task type largely determine how severely a model degrades.
To understand how these factors shape degradation, we conduct systematic comparisons across bit‑widths, quantization methods, model scales and task types under a controlled workflow. To this end, we develop COSQuant\footnote{\url{https://github.com/Cfish808/LLM-Compression-Tool}}, a unified quantization and evaluation toolkit. Using the standardized weight-only PTQ pipeline, we conduct over 800 evaluations covering 76 quantized configurations, 5 model families, 4 quantization methods, 4 bit-widths and 11 downstream benchmarks. Across these configurations, degradation changes markedly with precision: 4-bit quantization typically retains performance, while 2-bit quantization often causes broad performance degradation. At 3-bit, the degradation becomes evident but highly variable across settings, with severity depending strongly on model scale, quantization method, and task type.


This variability raises a key mechanistic question: why do models quantized to the same bit‑width degrade to different extents across model scales, methods and tasks? Conventional metrics such as perplexity and downstream accuracy can identify where degradation appears, but they do not explain why it varies across these settings~\cite{li2025quantization,zhou2026signal}. Answering this question requires comparing the internal computation of the full‑precision and quantized models, so that observed degradation can be traced to the errors introduced by quantization.

A signal‑to‑noise perspective provides a natural way to make this comparison. Intuitively, the impact of quantization error depends not only on its magnitude, but also on the strength of the signal it disrupts: the same amount of error can be harmless when the signal is strong or redundant, but disruptive when the signal is weak or task-critical. In this view, the signal is the full-precision representation, and the noise is the error introduced by low-precision approximation. We therefore use the signal-to-noise ratio (SNR) to measure the strength of the full-precision signal relative to the quantization noise. Comparing SNR across controlled settings allows us to examine whether differences in degradation correspond to how strongly quantization perturbs internal representations. This aligns with recent work that uses SNR to characterize the internal fidelity of compressed models~\cite{zhou2025revisiting,federici2026dissecting}. Through extensive controlled comparisons, we verify that SNR reliably captures downstream performance degradation, establishing it as an informative internal metric for interpreting quantization degradation.

Guided by this signal-to-noise perspective, we conduct two complementary analyses that together reveal how quantization degrades model performance. First, we measure how errors are introduced: for each quantized module, we derive an SNR decomposition that separates three components: the magnitude of the weight quantization error, the strength of the signal under a specific task, and the alignment between the error and task-specific activations. This decomposition shows that different factors affect these components in distinct ways, while model‑scale effects are coupled across all components rather than dominated by any single one. Second, we track how these errors propagate: we compare the internal representations of full‑precision models with those of their quantized counterparts across depth, measuring how errors accumulate through subsequent layers. This analysis reveals that larger models benefit not only from smaller errors introduced at each layer but also from weaker amplification as errors propagate. Together, these analyses reframe quantization degradation from an external performance decline into an internal account of where errors arise and how they accumulate across layers.

In summary, this work provides a systematic account of quantization degradation by combining large-scale empirical evaluation with signal-to-noise-based internal analysis. Our experiments show that degradation severity varies substantially with model scale, quantization method, and task type, rather than being determined by bit-width alone. Using SNR as an internal metric, we trace this variability back to two linked processes: how quantization error arises within individual modules, and how it accumulates across layers. This perspective suggests that reliable quantization should not be evaluated only by average performance retention. It also requires understanding what shapes degradation severity, and which internal processes drive it.

\section{Results}

\subsection{Bit-width alone does not determine quantization degradation}
\label{sec:results_degradation}

\definecolor{qdbvBlue}{HTML}{8FB8D2}
\providecommand{\qdbvHcell}[2]{%
  \begingroup
  \setlength{\fboxsep}{0.7pt}%
  \colorbox{qdbvBlue!#1}{\makebox[3.25em][c]{\strut #2}}%
  \endgroup
}

\begin{table}[!htbp]
\centering
\setlength{\belowcaptionskip}{2pt}
\setlength{\tabcolsep}{2.7pt}
\renewcommand{\arraystretch}{1.00}
\caption{
    \textbf{Detailed downstream retention for the degradation analysis.}
    Retention is computed over the accuracy-based downstream tasks described in Section~\ref{sec:methods_quant_eval}, after de-randomization and grouping into three task types: commonsense, knowledge, and reasoning.
    \textbf{a}, Method-dependent retention across model families and task types. Overall retention averages the three task types; W4$\rightarrow$W3 reports the retention decrease in percentage points. The last three columns report W3 retention within each task type.
    \textbf{b}, Scale-dependent retention by bit-width and task type for Qwen3 models quantized with GPTQ. 
    Darker shading in \textbf{a} denotes stronger degradation (lower retention or a larger W4$\rightarrow$W3 decrease). Values slightly above 100\% can occur when the quantized model marginally outperforms its full-precision counterpart, as is sometimes observed at high precision (W8).
}
\label{tab:detailed_retention}

\begin{tabular*}{0.95\textwidth}{@{\extracolsep{\fill}}llcccccc@{}}
\multicolumn{8}{@{}l}{\textbf{a. Method-dependent retention across model families and task types.}} \\
\toprule
\multirow{2}{*}{Model} & \multirow{2}{*}{Method}
& \multicolumn{3}{c}{Overall retention (\%)}
& \multicolumn{3}{c}{W3 retention (\%)} \\
\cmidrule(lr){3-5}\cmidrule(lr){6-8}
& & W4 & W3 & W4$\rightarrow$W3 & Comm. & Know. & Reas. \\
\midrule
\multirow{4}{*}{Llama-2-7B} & RTN & \qdbvHcell{4}{95.1} & \qdbvHcell{34}{60.0} & \qdbvHcell{50}{35.1} & \qdbvHcell{15}{82.0} & \qdbvHcell{47}{44.3} & \qdbvHcell{39}{53.7} \\
 & AWQ & \qdbvHcell{6}{92.8} & \qdbvHcell{23}{72.6} & \qdbvHcell{29}{20.2} & \qdbvHcell{9}{89.5} & \qdbvHcell{28}{67.5} & \qdbvHcell{33}{60.9} \\
 & GPTQ & \qdbvHcell{5}{94.3} & \qdbvHcell{16}{81.6} & \qdbvHcell{18}{12.7} & \qdbvHcell{5}{93.8} & \qdbvHcell{22}{74.6} & \qdbvHcell{20}{76.4} \\
 & SPQR & \qdbvHcell{4}{94.9} & \qdbvHcell{9}{89.2} & \qdbvHcell{8}{5.7} & \qdbvHcell{4}{95.2} & \qdbvHcell{13}{85.2} & \qdbvHcell{11}{87.2} \\
\midrule
\multirow{4}{*}{Llama-2-13B} & RTN & \qdbvHcell{4}{95.1} & \qdbvHcell{21}{75.7} & \qdbvHcell{27}{19.4} & \qdbvHcell{10}{88.6} & \qdbvHcell{19}{77.1} & \qdbvHcell{33}{61.5} \\
 & AWQ & \qdbvHcell{8}{90.1} & \qdbvHcell{14}{83.5} & \qdbvHcell{9}{6.6} & \qdbvHcell{7}{91.8} & \qdbvHcell{11}{86.8} & \qdbvHcell{24}{71.9} \\
 & GPTQ & \qdbvHcell{3}{96.5} & \qdbvHcell{11}{87.3} & \qdbvHcell{13}{9.2} & \qdbvHcell{4}{95.5} & \qdbvHcell{12}{85.5} & \qdbvHcell{16}{80.8} \\
 & SPQR & \qdbvHcell{2}{97.5} & \qdbvHcell{7}{91.6} & \qdbvHcell{8}{5.9} & \qdbvHcell{2}{97.6} & \qdbvHcell{5}{93.7} & \qdbvHcell{14}{83.5} \\
\midrule
\multirow{4}{*}{Llama-3-8B} & RTN & \qdbvHcell{7}{91.6} & \qdbvHcell{52}{39.3} & \qdbvHcell{74}{52.3} & \qdbvHcell{34}{60.2} & \qdbvHcell{57}{33.5} & \qdbvHcell{65}{24.1} \\
 & AWQ & \qdbvHcell{8}{90.3} & \qdbvHcell{23}{73.4} & \qdbvHcell{24}{16.9} & \qdbvHcell{13}{84.9} & \qdbvHcell{24}{71.2} & \qdbvHcell{30}{64.2} \\
 & GPTQ & \qdbvHcell{5}{93.6} & \qdbvHcell{28}{67.5} & \qdbvHcell{37}{26.1} & \qdbvHcell{13}{84.7} & \qdbvHcell{36}{57.9} & \qdbvHcell{34}{59.8} \\
 & SPQR & \qdbvHcell{2}{98.1} & \qdbvHcell{10}{88.5} & \qdbvHcell{14}{9.6} & \qdbvHcell{5}{93.6} & \qdbvHcell{12}{85.5} & \qdbvHcell{12}{86.4} \\
\midrule
\multirow{4}{*}{Qwen2.5-7B} & RTN & \qdbvHcell{5}{94.2} & \qdbvHcell{51}{40.0} & \qdbvHcell{77}{54.2} & \qdbvHcell{44}{48.5} & \qdbvHcell{51}{40.0} & \qdbvHcell{58}{31.6} \\
 & AWQ & \qdbvHcell{11}{86.7} & \qdbvHcell{14}{83.9} & \qdbvHcell{4}{2.8} & \qdbvHcell{10}{88.5} & \qdbvHcell{14}{83.0} & \qdbvHcell{17}{80.1} \\
 & GPTQ & \qdbvHcell{3}{96.5} & \qdbvHcell{15}{82.3} & \qdbvHcell{20}{14.2} & \qdbvHcell{8}{90.2} & \qdbvHcell{14}{83.4} & \qdbvHcell{23}{73.3} \\
 & SPQR & \qdbvHcell{2}{97.1} & \qdbvHcell{7}{91.6} & \qdbvHcell{8}{5.5} & \qdbvHcell{3}{96.1} & \qdbvHcell{8}{90.6} & \qdbvHcell{10}{88.2} \\
\midrule
\multirow{4}{*}{Qwen3-8B} & RTN & \qdbvHcell{6}{92.5} & \qdbvHcell{53}{37.9} & \qdbvHcell{77}{54.6} & \qdbvHcell{36}{58.1} & \qdbvHcell{58}{32.3} & \qdbvHcell{65}{23.3} \\
 & AWQ & \qdbvHcell{4}{95.2} & \qdbvHcell{15}{82.7} & \qdbvHcell{18}{12.5} & \qdbvHcell{9}{89.9} & \qdbvHcell{18}{78.9} & \qdbvHcell{18}{79.3} \\
 & GPTQ & \qdbvHcell{2}{97.4} & \qdbvHcell{14}{83.3} & \qdbvHcell{20}{14.1} & \qdbvHcell{7}{91.5} & \qdbvHcell{18}{79.0} & \qdbvHcell{17}{79.5} \\
 & SPQR & \qdbvHcell{2}{97.3} & \qdbvHcell{7}{91.2} & \qdbvHcell{9}{6.1} & \qdbvHcell{2}{97.1} & \qdbvHcell{9}{89.0} & \qdbvHcell{11}{87.6} \\
\midrule
\multirow{4}{*}{Qwen3-14B} & RTN & \qdbvHcell{3}{96.9} & \qdbvHcell{31}{64.0} & \qdbvHcell{47}{32.9} & \qdbvHcell{18}{78.4} & \qdbvHcell{35}{58.9} & \qdbvHcell{39}{54.7} \\
 & AWQ & \qdbvHcell{3}{96.1} & \qdbvHcell{11}{87.1} & \qdbvHcell{13}{9.0} & \qdbvHcell{5}{93.7} & \qdbvHcell{13}{84.8} & \qdbvHcell{15}{82.8} \\
 & GPTQ & \qdbvHcell{2}{98.1} & \qdbvHcell{8}{90.8} & \qdbvHcell{10}{7.3} & \qdbvHcell{4}{95.5} & \qdbvHcell{11}{87.6} & \qdbvHcell{9}{89.1} \\
 & SPQR & \qdbvHcell{1}{98.7} & \qdbvHcell{5}{94.5} & \qdbvHcell{6}{4.2} & \qdbvHcell{2}{97.4} & \qdbvHcell{7}{92.3} & \qdbvHcell{5}{93.9} \\
\midrule
Mistral-7B & GPTQ & \qdbvHcell{2}{97.4} & \qdbvHcell{9}{89.1} & \qdbvHcell{12}{8.3} & \qdbvHcell{3}{96.4} & \qdbvHcell{13}{84.9} & \qdbvHcell{12}{86.1} \\
\bottomrule
\end{tabular*}

\vspace{1.0em}

\begin{tabular*}{0.95\textwidth}{@{\extracolsep{\fill}}llccccc@{}}
\multicolumn{7}{@{}l}{\textbf{b. Scale-dependent retention by bit-width and task type.}} \\
\addlinespace[0.25em]
\toprule
Bit-width & Metric & Qwen3-0.6B & Qwen3-1.7B & Qwen3-4B & Qwen3-8B & Qwen3-14B \\
\midrule
W8 & Overall & 100.7 & 100.8 & 100.1 & 100.2 & 100.1 \\
W4 & Overall & 70.8 & 80.7 & 95.4 & 95.9 & 98.1 \\
\addlinespace[0.15em]
\midrule
\multirow{4}{*}{W3} & Overall & 30.6 & 44.2 & 72.4 & 81.2 & 87.9 \\
 & Commonsense & 55.3 & 70.2 & 85.2 & 89.5 & 93.7 \\
 & Knowledge & 15.5 & 39.0 & 66.3 & 77.4 & 83.6 \\
 & Reasoning & 20.9 & 23.5 & 65.7 & 76.5 & 86.4 \\
\midrule
W2 & Overall & 6.3 & 3.6 & 2.5 & 2.2 & 8.4 \\
\bottomrule
\end{tabular*}
\end{table}

\begin{figure}[!t]
    \centering
    \includegraphics[width=\linewidth]{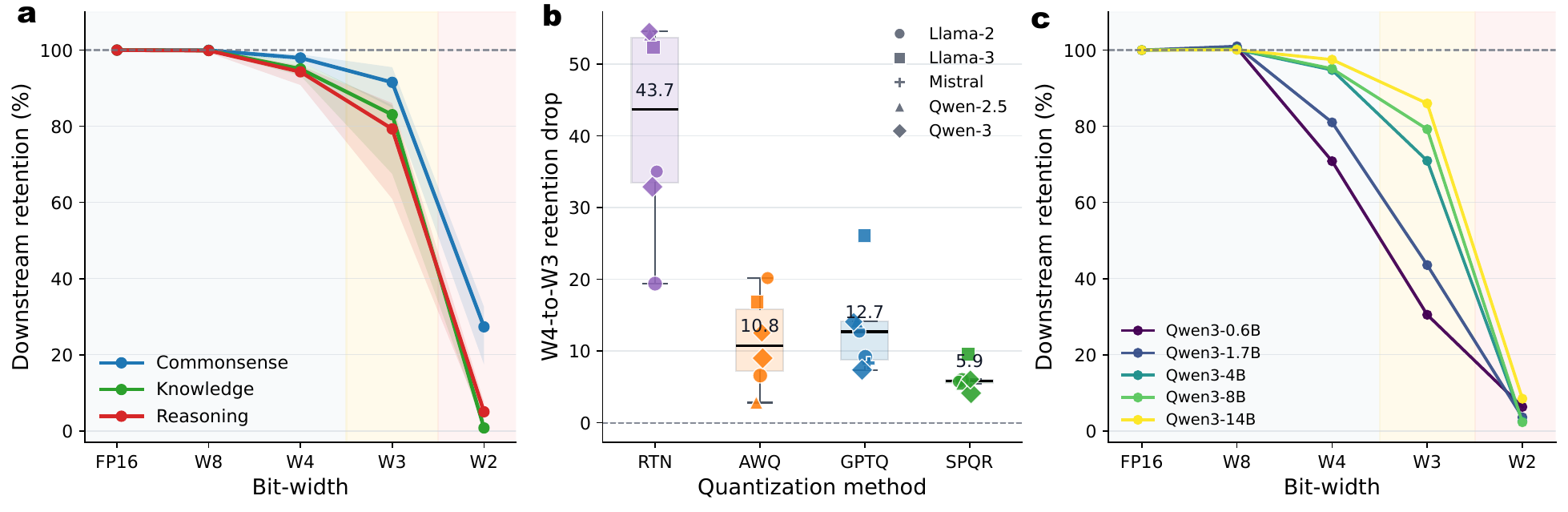}
    \caption{
        \textbf{Quantization degradation varies with task type, quantization method, and model scale.}
        \textbf{a}, Downstream retention by task type across bit-widths, summarized across models at 7B–14B scale and all quantization methods studied. Lines show median within each task type; shaded areas indicate the range between the 25th and 75th percentiles across models and quantization methods.
        \textbf{b}, Drop in downstream retention from W4 to W3, averaged across task types with equal weight. Results are shown for each quantization method. Each point denotes one model setting, and boxes summarize the distribution within each method.
        \textbf{c}, Qwen3 scaling analysis under GPTQ, showing downstream retention across bit-widths, averaged across task types with equal weight.
        Results for individual downstream benchmarks and the Qwen3 scaling curves for each task type are shown in Supplementary Figs.~\ref{fig:app_task_level_retention} and~\ref{fig:app_scale_retention_by_task_type}, respectively.
    }
    \label{fig:degradation-boundaries}
\end{figure}

Even at the same bit-width, models can range from high retention to severe degradation, depending on model scale, quantization method, and task type. To quantify this variation, we systematically evaluate 76 quantized configurations across 5 model families, 4 quantization methods, 4 bit-widths, and 11 downstream benchmarks. We report de‑randomized downstream retention, which subtracts task‑specific random baselines before computation. The corresponding numerical results are summarized in Table~\ref{tab:detailed_retention}.

Figure~\ref{fig:degradation-boundaries}a shows how degradation varies with task type across bit-widths. From FP16 to W4, performance remains largely preserved across all task types. At W3, clear differences emerge: commonsense reasoning tasks retain substantially more of their performance, whereas knowledge-intensive and reasoning tasks degrade more sharply. This gap persists across model families and quantization methods (Table~\ref{tab:detailed_retention}a). At W2, degradation becomes severe across all task types, with knowledge and reasoning performance approaching near-collapse. Thus, W3 is the precision at which task-type differences become most pronounced.

These differences are not limited to task type. Figure~\ref{fig:degradation-boundaries}b examines how the choice of quantization method affects degradation at the W4-to-W3 transition, where performance gaps are largest. Among the methods evaluated, RTN exhibits the largest median drop (43.7 percentage points), indicating that naive rounding is most vulnerable at this transition. AWQ and GPTQ show substantially smaller median drops (10.8 and 12.7, respectively), while SPQR exhibits the smallest drop (5.9 percentage points). In short, the choice of quantization method strongly shapes how severely a model degrades.

Model scale affects degradation as well. In the Qwen3 scaling analysis (Fig.~\ref{fig:degradation-boundaries}c), smaller models degrade earlier as precision is reduced, whereas larger models, particularly the 8B and 14B models, remain more stable at W4 and W3. Table~\ref{tab:detailed_retention}b further shows how this scale effect differs by task type: on commonsense reasoning tasks, even relatively small models retain a high fraction of performance at W3, whereas knowledge and reasoning tasks require substantially larger models to achieve similarly high retention. However, this scale advantage disappears at W2, where all tested model scales degrade severely. Larger models can therefore postpone the onset of severe degradation, but they cannot prevent it under extreme quantization.

Together, these results show that the same bit-width can lead to markedly different degrees of degradation depending on the task being evaluated, the quantization method used, and the model scale. This empirical observation motivates our subsequent analysis of the internal mechanisms behind these shifts.

\subsection{SNR captures relative performance retention}
\label{sec:results_pairwise_snr}

\begin{figure}[!t]
    \centering
    \begin{minipage}[t]{0.36\textwidth}
        \centering
        \includegraphics[width=\linewidth]{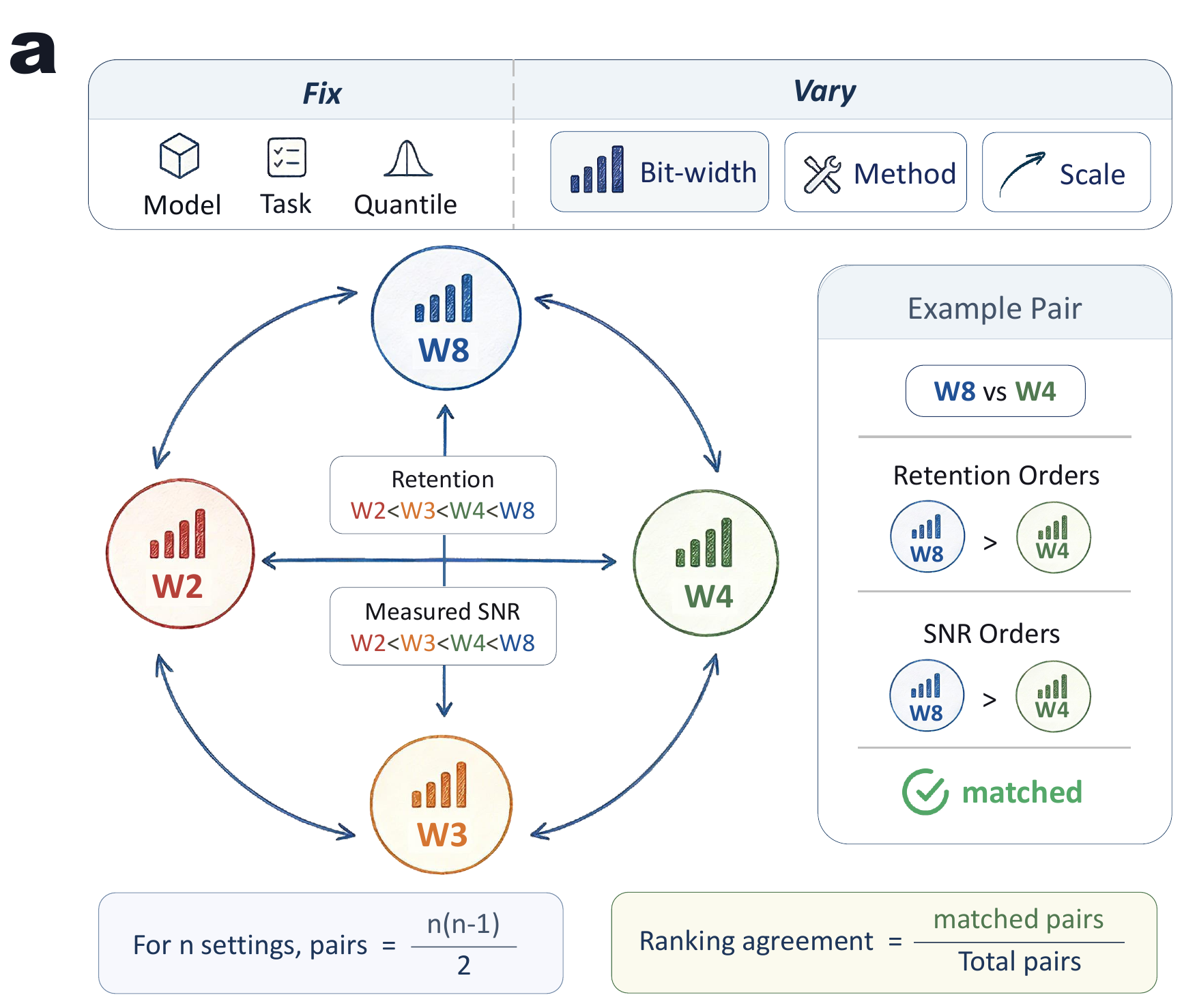}
    \end{minipage}
    \hfill
    \begin{minipage}[t]{0.63\textwidth}
        \centering
        \includegraphics[width=\linewidth]{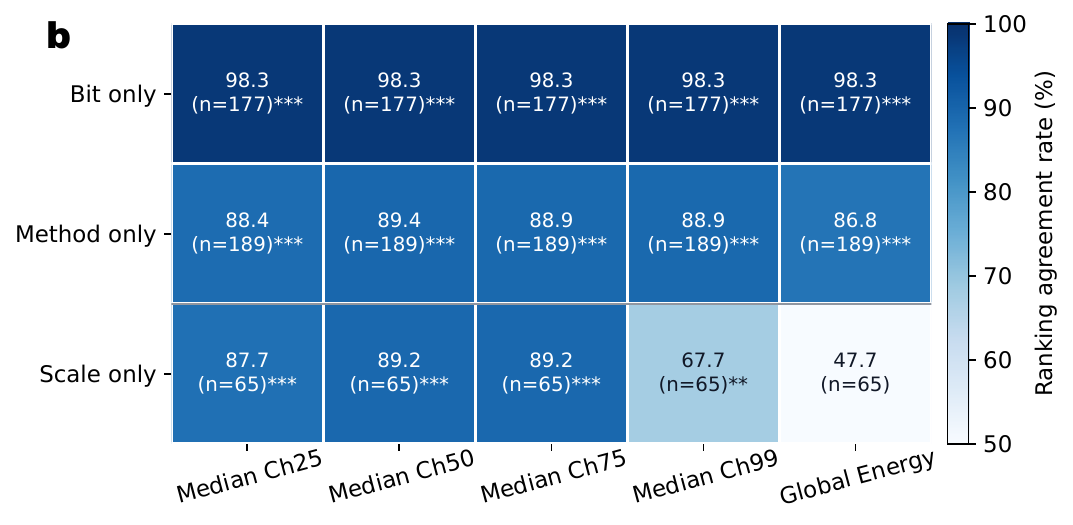}
    \end{minipage}

    \caption{
        \textbf{SNR captures relative performance retention in controlled pairwise comparisons.}
        \textbf{a}, Schematic of the pairwise comparison. In each comparison, a single factor (bit-width, quantization method, or model scale) is varied while all other factors are fixed. Every possible pair within a group is evaluated, and a pair is counted as agreeing when the ordering by SNR matches the ordering by performance retention.
        \textbf{b}, Agreement rates between SNR and performance retention when varying bit-width, quantization method, or model scale. Scale comparisons are formed within the same model family. Columns correspond to different ways of summarizing SNR across layers and channels. Median Ch\(x\) denotes the across-layer median of the \(x\)-th percentile of channel-wise SNR. Global Energy is the ratio of total full-precision representation energy to total quantization error energy. \(n\) denotes the number of valid pairs after excluding pairs with near-identical retention. Stars mark agreement rates significantly above the \(50\%\) chance level under a one-sided binomial test.
    }
    \label{fig:pairwise_comparison}
\end{figure}

The preceding analysis shows that quantization can lead to markedly different performance degradation across bit-widths, quantization methods, model scales, and task types. We now ask whether this variability in external evaluation scores is also reflected in the model's internal representations. Following the signal-to-noise perspective established earlier, we treat the full-precision representation as signal and the quantization-induced approximation error as noise. The signal-to-noise ratio (SNR) therefore measures the strength of the signal relative to the quantization noise, with higher SNR indicating better preservation of the original signal (Eq.~\eqref{eq:channel_snr}). This makes SNR a natural internal measure of representation fidelity, consistent with recent analyses of compressed language models~\cite{zhou2025revisiting}.

\begin{figure}[!t]
    \centering
    \includegraphics[width=1.0\textwidth]{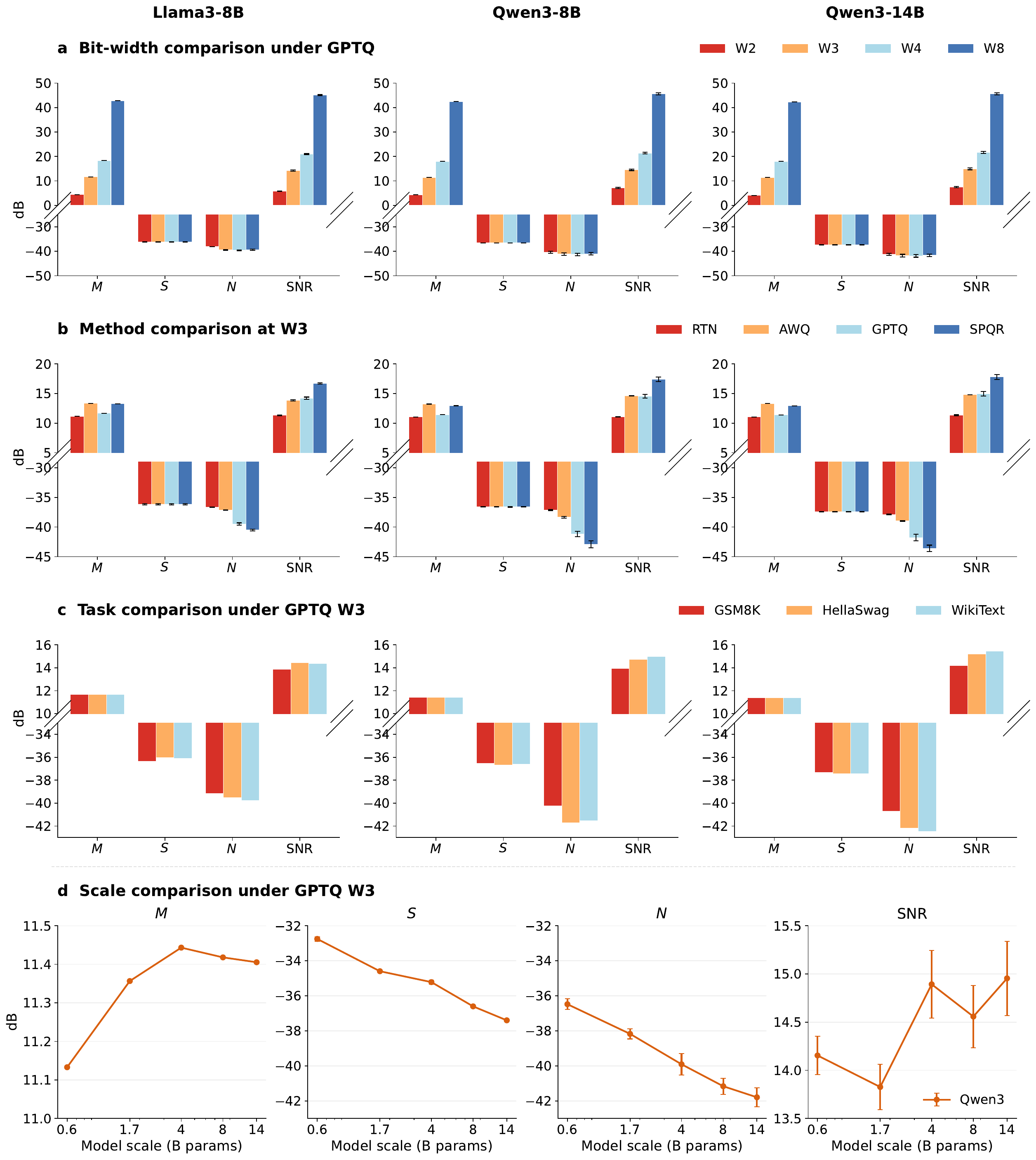}
    \caption{
        \textbf{Source SNR decomposes into magnitude, signal, and noise components.}
        \textbf{a}, Bit-width effect under GPTQ. Reducing precision primarily lowers the magnitude term \(M\).
        \textbf{b}, Method effect at W3. Quantization methods differ not only in \(M\), but also in the noise term \(N\).
        \textbf{c}, Task effect under GPTQ W3. \(M\) is unchanged across tasks, while differences in \(S\) and especially \(N\) alter the source SNR.
        \textbf{d}, Scale comparison under GPTQ-W3. Model scale changes all three components simultaneously, rather than through a single dominant one.
        All panels report channel-median statistics in decibels (dB). Values are first aggregated by taking the median across layers within each module and then across modules. Points denote group means, with error bars indicating standard errors.
        Results for individual linear modules and sensitivity to alternative channel and layer summaries are reported in Supplementary Fig.~\ref{fig:app_modulewise_source_snr} and Supplementary Table~\ref{tab:app_summary_statistic_robustness}, respectively.
    }    
    \label{fig:msn}
\end{figure}

We use the controlled pairwise comparison illustrated in Fig.~\ref{fig:pairwise_comparison}a. In each comparison, a single factor is varied while the others are fixed. For example, bit-width comparisons vary only the precision, keeping the model, method and task identical. Within each such controlled group, all possible pairs of settings are formed. A pair is counted as agreeing when the setting with higher performance retention also yields higher SNR. This ordinal design does not require predicting absolute retention scores. The factors examined are bit-width, quantization method, and model scale; task-dependent effects are examined separately (Fig.~\ref{fig:msn}c). 

Bit-width comparisons show the strongest agreement (Fig.~\ref{fig:pairwise_comparison}b). Across all SNR summaries, the agreement rate reaches \(98.3\%\) (\(n=177\)), indicating that settings differing only in bit-width are almost always ordered consistently by SNR and performance retention. Method comparisons also show high agreement, ranging from \(86.8\%\) to \(89.4\%\) (\(n=189\)), demonstrating that SNR captures performance differences across quantization methods.
Scale comparisons are more sensitive to how SNR is summarized. Comparisons achieve \(87.7\%\)–\(89.2\%\) agreement with typical-channel statistics (Median Ch25, Ch50, and Ch75), but drop to \(67.7\%\) for a tail-channel statistic (Median Ch99) and \(47.7\%\) for Global Energy. The contrast shows that the choice of summary statistic matters for scale comparisons. Typical-channel summaries capture the behavior of the majority of channels across layers, whereas Median Ch99 focuses on the highest-SNR channels and Global Energy combines all layers and channels into a single total-energy ratio. This motivates us to adopt typical-channel summaries in subsequent analyses, rather than relying solely on Global Energy.

Together, these results support SNR as a relative internal measure of quantization-induced degradation. The alignment is strongest for direct quantization factors such as bit-width and method, and remains informative for model scale when SNR is summarized over typical channels. 

\subsection{Source SNR is shaped by magnitude, signal and noise effects}
\label{sec:results_msn}

We next examine what shapes the SNR at the position where quantization error first enters the model. For each quantized module, we measure its output SNR using clean activations from the full-precision model. We refer to this measure as the source SNR. To understand what drives this error, we decompose the source SNR for each output channel \(c\) at layer \(l\) into three components. Expressed in decibels as the $10\log_{10}$ ratio of signal energy to error energy, this ratio decomposes additively:
\begin{equation}
    \mathrm{SNR}_{l,c}^{\mathrm{source}} = M_{l,c}+S_{l,c}-N_{l,c},
\end{equation}
with the mathematical derivation and definitions provided in Section~\ref{sec:methods_msn}. \(M\) measures the magnitude of the original weight relative to its quantization error, with larger values indicating a smaller relative error. \(S\) measures the energy of the task-specific activations along the direction of the full-precision weight, whereas \(N\) measures the activation energy along the direction of quantization error. A larger \(S\) therefore corresponds to a stronger full-precision signal, while a larger \(N\) indicates that the quantization error is more strongly expressed under the task's activations. Larger \(M\) and \(S\) increase the source SNR, whereas larger \(N\) decreases it. This decomposition distinguishes whether degradation arises from a larger relative weight error, a weaker signal, or stronger alignment between the error and task-specific activations.

Figure~\ref{fig:msn} examines how bit-width, quantization method, downstream task, and model scale each affect the three source components. Bit-width primarily controls the size of the weight quantization error. Under GPTQ, reducing precision from W8 to W2 produces a steady decline in \(M\), and the source SNR falls accordingly (Fig.~\ref{fig:msn}a). The noise term \(N\) varies only slightly across bit-widths, and the signal term \(S\) remains constant because it depends solely on the original weights and task activations, neither of which changes when precision alone is varied. Consequently, the decline in source SNR at lower precision is driven primarily by the decrease in \(M\), which reflects larger quantization error relative to the original weight.

Quantization methods affect both the size of the weight quantization error and its alignment with task-specific activations. At W3, SPQR achieves the highest source SNR across representative models (Fig.~\ref{fig:msn}b). Its advantage comes from both a higher \(M\) and a lower \(N\). AWQ also shows a competitive \(M\), but its larger \(N\) partially offsets this advantage. GPTQ is limited primarily by a lower \(M\), whereas RTN suffers from both a low \(M\) and a high \(N\). Method performance therefore depends not only on how much error is introduced, but also on how strongly that error aligns with task-specific activations.

Task effects are mediated by the activation-dependent terms. We compare WikiText-2~\cite{merity2016pointera}, HellaSwag~\cite{zellers2019hellaswag} and GSM8K~\cite{cobbe2021training} as representative tasks spanning three distinct task types: language modeling, commonsense reasoning, and mathematical reasoning. These tasks elicit markedly different activation statistics, allowing us to probe how the same quantized model behaves under different task demands. For the same model, method, and bit-width, the magnitude term \(M\) is unchanged across tasks because both the weights and their quantization errors are fixed (Fig.~\ref{fig:msn}c). All task-driven variation in source SNR therefore comes from \(S\) and \(N\). GSM8K exhibits the lowest source SNR among the three tasks, driven primarily by its larger \(N\), reflecting stronger alignment between the quantization error and task-specific activations.

Model scale couples all three components. Under GPTQ-W3, increasing the scale of Qwen3 models changes \(M\), \(S\), and \(N\) simultaneously, rather than through a single dominant component (Fig.~\ref{fig:msn}d). Specifically, \(M\) increases from the smallest to mid-scale models and then declines slightly. \(S\) and \(N\) both tend to decrease with scale, but they have opposing effects on the source SNR: a lower \(S\) weakens the signal response, whereas a lower \(N\) reduces the noise. Combined with the increase in \(M\), this yields a higher source SNR for larger models, although the trend is not strictly monotonic. The scale effect is inherently coupled, and cannot be reduced to a single factor.

Taken together, these results show that the factors shaping degradation affect the source SNR in distinct ways. Bit-width acts primarily on the magnitude of the weight quantization error, quantization method additionally shapes how the error aligns with task-specific activations, and task alters the signal and noise responses through the same activations. Model scale couples all three components, rather than operating through any single one. This decomposition provides the first part of the signal-to-noise analysis: it identifies how quantization error is generated before it propagates across layers.

\clearpage

\begin{figure}[!t]
    \centering
    \includegraphics[width=1.0\textwidth]{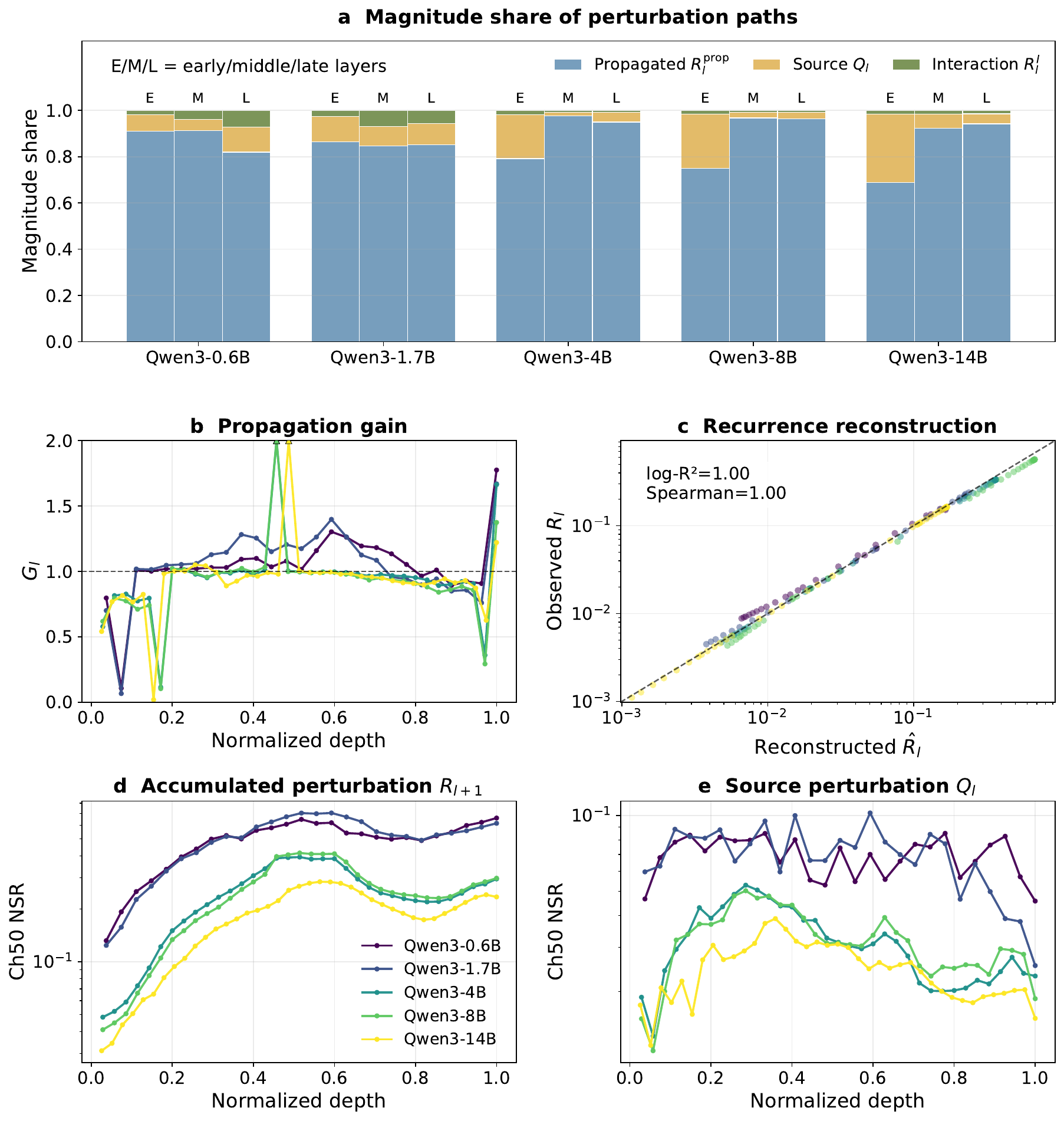}
    \caption{
        \textbf{Cross-layer decomposition of accumulated NSR.} Qwen3 models from 0.6B to 14B are quantized with GPTQ-W3 and analyzed on WikiText-2.
        \textbf{a}, Relative contributions of propagated error \(R_l^{\mathrm{prop}}\), source error \(Q_l\), and their interaction \(R_l^I\) to the total accumulated error. The three stages correspond to equal thirds of the model's total depth.
        \textbf{b}, Propagation gain \(G_l\) across layers. The dashed line marks \(G_l = 1\); values outside the plotted range are clipped for readability.
        \textbf{c}, Reconstructed versus observed accumulated NSR. The reconstructed \(\widehat{R}_l\) is obtained by iterating the recurrence using the measured \(G_l\) and \(Q_l\).
        \textbf{d}, Accumulated NSR \(R_{l+1}\) across layers (Ch50).
        \textbf{e}, Source NSR \(Q_l\) across layers (Ch50).
        Panels \textbf{a}--\textbf{c} use global Frobenius-energy ratios, whereas panels \textbf{d} and \textbf{e} report Ch50 NSR, computed as the median over channels at each layer. Corresponding analyses on HellaSwag and GSM8K are shown in Supplementary Figs.~\ref{fig:app_source_propagation_hellaswag} and~\ref{fig:app_source_propagation_gsm8k}.
    }
    \label{fig:source_propagation}
\end{figure}

\subsection{Cross-layer propagation shapes how quantization errors accumulate}
\label{sec:results_source_propagation}



To separate the error introduced at each layer from the error accumulated through propagation, we switch from SNR to the noise-to-signal ratio (NSR), the inverse energy form of SNR. Higher NSR indicates larger error relative to the full-precision signal, making it suitable for decomposing errors layer by layer. The cross‑layer accumulation can be summarized by the recurrence
\begin{equation}
    R_{l+1}\approx G_l R_l + Q_l ,
    \label{eq:source_propagation_recurrence}
\end{equation}
where \(R_l\) is the accumulated error entering block \(l\), and \(Q_l\) is the error newly introduced by the current block. The factor \(G_l\) represents the propagation gain, which measures how the accumulated error is altered after passing through the current block. Specifically, \(G_l < 1\) indicates error attenuation, \(G_l > 1\) indicates error amplification, and \(G_l \approx 1\) means that the error passes through largely unchanged. Both \(R_l\) and \(Q_l\) are expressed as NSR, so that larger values indicate larger error relative to the full-precision signal. 

This recurrence separates the total error after each block into the error propagated from earlier layers (\(G_lR_l\)) and the error newly introduced at the current block (\(Q_l\)), referred to below as propagated error and source error. The formal definitions, including the residual interaction term \(R_l^I\) omitted from this reduced recurrence, are provided in Section~\ref{sec:methods_source_propagation}.

We apply this analysis to model scale because larger models differ not only in the errors they introduce, but also in the depth and width through which those errors propagate. The source SNR analysis already showed that scale couples \(M\), \(S\), and \(N\), and that the resulting SNR does not always increase with scale, suggesting that the effect of model scale cannot be fully understood by local analysis alone.

Figure~\ref{fig:source_propagation} examines this accumulation process across Qwen3 model scales by addressing three questions. First, where does the accumulated error at different depths come from? We separate the contributions of propagated error, source error, and their interaction (Fig.~\ref{fig:source_propagation}a). Second, can the accumulated NSR be reconstructed through the recurrence? We measure the propagation gain \(G_l\) and test whether the observed accumulation can be reproduced by iterating Eq.~\eqref{eq:source_propagation_recurrence} from the first layer (Fig.~\ref{fig:source_propagation}b,c). Finally, how does model scale alter this process? We track source and accumulated errors across depth to compare how errors are introduced and propagated at different scales (Fig.~\ref{fig:source_propagation}d,e).

Fig.~\ref{fig:source_propagation}a shows that the propagated error \(R_l^{\mathrm{prop}}\) (i.e., \(G_lR_l\)) dominates the total error across early, middle and late stages. While the source error \(Q_l\) remains noticeable in early layers, it becomes secondary as errors build up through depth. Meanwhile, the interaction term \(R_l^I\), which captures the residual not explained by the propagation and source terms alone, remains comparatively small across most stages. This dominance shows why analyzing source errors alone is insufficient: at later layers, degradation is driven less by the errors newly introduced there than by how accumulated errors are propagated.

Fig.~\ref{fig:source_propagation}b tracks the propagation gain \(G_l\) across layers to examine how the accumulated error evolves as it passes through each block. Clear differences emerge across model scales in the middle layers: smaller Qwen3 models exhibit sustained amplification, whereas larger models largely preserve or mildly attenuate the error. While occasional layers in larger models show pronounced amplification peaks, they are isolated and do not reflect the overall trend. These curves reveal that accumulated error is not simply the sum of source errors introduced at each layer; rather, each block can either attenuate or amplify the error it receives. Using the measured \(G_l\) and \(Q_l\) in the recurrence of Eq.~\eqref{eq:source_propagation_recurrence}, the reconstructed \(\widehat{R}_l\) closely matches the observed accumulated NSR (Fig.~\ref{fig:source_propagation}c). This agreement indicates that \(G_l\) and \(Q_l\) capture the essential dynamics of error accumulation, without requiring the residual interaction term.

The layer-wise NSR curves in Fig.~\ref{fig:source_propagation}d,e show how model scale affects both source and accumulated errors. Larger Qwen3 models consistently exhibit both lower source error \(Q_l\) and accumulated error \(R_{l+1}\) across most layers. While this trend is not strictly followed by adjacent scales, the overall trend reveals two complementary scale advantages: larger models introduce smaller source errors and avoid the sustained amplification observed in smaller models. These two effects jointly account for why larger models maintain lower accumulated NSR after quantization.

Together, these findings extend the signal‑to‑noise analysis from source errors to cross‑layer accumulation. The M/S/N decomposition identifies how source errors are introduced at each module, whereas the accumulated NSR analysis tracks how those errors are attenuated, amplified, or passed through unchanged across depth. This distinction underscores that degradation depends not only on where errors are originally generated, but also on how those errors evolve as they pass across layers.

\section{Discussion}



\noindent\textbf{Rethinking quantization evaluation}\par
\noindent
Our results show that models compressed to the same bit-width can exhibit substantially different degradation depending on the task being evaluated, the quantization method used, and the model scale. This variability means that bit-width alone is insufficient to determine how severely a model degrades. Evaluation should therefore move beyond simply asking whether a given bit-width is generally acceptable, and instead ask whether a setting that performs well at one precision may degrade substantially for certain tasks, methods, or scales. 

Metrics such as perplexity and average downstream retention can mask this variation: a configuration that appears acceptable on average may already degrade sharply for knowledge-intensive or reasoning tasks. A more informative approach is to identify where these differences are most pronounced. At 4-bit, performance is largely retained, whereas 2-bit often leads to broad degradation. Both obscure the distinct influences of quantization method, task type, and model scale. At 3-bit, by contrast, performance differences across these factors become most pronounced. This provides a clear evaluation insight: future evaluations should specifically target such stress points, rather than relying solely on precision levels where performance is either largely retained or already broadly degraded.

\noindent\textbf{How errors are introduced and propagated shapes degradation}\par
\noindent
To understand why degradation varies substantially across settings, we turn to the signal–noise perspective. The M/S/N decomposition shows that each factor affects the error introduced by quantized modules in distinct ways: bit-width primarily controls the magnitude of the weight quantization error, quantization method additionally shapes how the error aligns with task-specific activations, and task alters the activation statistics, so that the same quantized weights produce different errors depending on the task. This is why the same precision can lead to different degradation at the source.

However, how errors are introduced does not fully account for how a model degrades. Once introduced, errors can be attenuated, amplified, or passed through largely unchanged as they propagate. This is especially important for model scale: larger models benefit not only from introducing less error at each layer, but also from weaker amplification during propagation. Degradation is therefore shaped by two linked processes: how errors arise under task-specific activations, and how they propagate across layers.

These two processes clarify why local reconstruction error alone is an incomplete measure of quantization quality. A small reconstruction error can still cause significant harm if it aligns strongly with task-specific activations or accumulates across layers. Conversely, a larger error may have limited impact if it is less aligned with these activations or is attenuated during propagation. 


\noindent\textbf{Implications for designing quantization methods}\par
\noindent
These findings suggest that future quantization methods should look beyond simply minimizing local reconstruction error. Most reconstruction-based objectives treat local error magnitude as the primary signal to optimize, but our results show that errors become harmful primarily when they align with task-specific activations, or when they persist and accumulate across layers. The optimization objective should therefore shift from uniformly minimizing all errors to selectively suppressing those that are both exposed by task activations and propagated across the network.

One implication concerns calibration data. Because calibration determines the activations seen during optimization, it directly influences which errors are identified as harmful. Generic calibration datasets may not capture the task‑specific activations that matter most for knowledge‑intensive or reasoning tasks. Task‑aware calibration can help target those errors more effectively. A second implication concerns mixed‑precision allocation: the layers or channels that deserve higher precision are not necessarily those with the largest local reconstruction error, but those whose errors are most strongly amplified by task activations or most prone to accumulate across layers. In both cases, the goal is to control the errors that drive downstream degradation, rather than to minimize all errors uniformly.

Our analysis focuses on weight-only post-training quantization, allowing weight-induced errors to be isolated and compared systematically across bit-widths, quantization methods, tasks and model scales. The quantitative findings should be interpreted within this scope, rather than directly extrapolated to activation quantization, quantization-aware training, or combined compression. However, the signal–noise framework itself is not tied to a specific quantization method. The mechanistic analyses developed here, including source SNR decomposition and cross-layer propagation analysis, are applicable whenever full-precision and compressed model representations can be compared. Future work could extend this framework to other compression paradigms and probe the causal role of source and accumulated errors in downstream degradation.

Taken together, our findings show that quantization degradation is shaped by two linked processes: how errors arise at individual modules, and how they propagate across layers. Reliable quantization therefore requires understanding both where errors originate and how they accumulate, rather than relying solely on average performance retention or local error minimization.

\section{Methods}

\subsection{Quantization and evaluation setup}
\label{sec:methods_quant_eval}

\paragraph{COSQuant toolkit.}
A systematic study of quantization degradation requires generating and evaluating multiple combinations of models, bit-widths, quantization methods, and tasks under a consistent workflow. We develop COSQuant, a unified and modular LLM compression toolkit that integrates calibration data preparation, model quantization and export, and benchmark evaluation through a configuration-driven pipeline. The toolkit supports PTQ, QAT, hybrid compression, and calibration-data optimization. For the present work, we rely on its standardized weight-only PTQ and evaluation workflow, which keeps quantization and downstream evaluation strictly comparable across all settings. The resulting paired full-precision and quantized models serve as the basis for both the degradation analysis and the signal-to-noise analysis.

\paragraph{Models, quantization methods, and bit-widths.}
We evaluate weight-only PTQ across several model families spanning multiple scales: Llama-2-7B and Llama-2-13B~\cite{touvron2023llama}, Llama-3-8B~\cite{grattafiori2024llama}, Mistral-7B~\cite{jiang2023mistral}, Qwen2.5-7B~\cite{qwen2025qwen25}, and Qwen3 models from 0.6B to 14B~\cite{yang2025qwen3}. The quantization methods examined are RTN, GPTQ~\cite{frantar2023iclr}, AWQ~\cite{lin2024mlsys} and SPQR~\cite{dettmers2024spqr}. Of these, GPTQ covers the broadest precision range, from W8 down to W2, while RTN, AWQ, and SPQR are evaluated at W4 and W3. In total, this yields 76 quantized configurations across 10 full‑precision models, enabling us to compare the four quantization methods at W4 and W3 and to isolate model-scale effects using the Qwen3 series.

\paragraph{Evaluation tasks.}
Downstream performance is assessed across three task types. Commonsense reasoning covers WinoGrande~\cite{sakaguchi2021winogrande}, HellaSwag~\cite{zellers2019hellaswag}, ARC-Easy~\cite{clark2018think}, PIQA~\cite{bisk2019piqa} and SocialIQA~\cite{sap2019socialiqa}. Knowledge-intensive tasks comprise TriviaQA~\cite{joshi2017triviaqa} and MMLU~\cite{hendrycks2021measuring}. Reasoning tasks span ARC-Challenge~\cite{clark2018think}, StrategyQA~\cite{geva2021did}, MathQA~\cite{amini2019mathqa} and GSM8K~\cite{cobbe2021training}.


\paragraph{Retention metrics and aggregation.}
For accuracy-based downstream tasks, we use a de-randomized retention score that compares each quantized model to its full-precision counterpart after removing the task-specific random baseline:
\begin{equation}
    \mathrm{Ret}_{\mathrm{acc}} =
    100\cdot
    \frac{\mathrm{Acc}_{\mathrm{q}}-\mathrm{Acc}_{\mathrm{rand}}}
    {\mathrm{Acc}_{\mathrm{fp}}-\mathrm{Acc}_{\mathrm{rand}}}.
\end{equation}
Here, the subscripts \(\mathrm{q}\) and \(\mathrm{fp}\) denote the quantized and full-precision models, respectively. For multiple-choice tasks, \(\mathrm{Acc}_{\mathrm{rand}}\) is determined by the number of answer choices, such as \(0.25\) for four-choice tasks and \(0.5\) for two-choice tasks. For open-ended exact-match tasks, the random baseline is treated as approximately zero.


For aggregation, retention scores are first averaged within each task type, then averaged across task types with equal weight to produce the overall downstream score. This weighting ensures that no single task type dominates the result. Results for individual downstream benchmarks are provided in Supplementary Fig.~\ref{fig:app_task_level_retention}.

\subsection{SNR and pairwise comparison}
\label{sec:methods_pairwise_snr}

\paragraph{SNR summaries.}
We use SNR to quantify how strongly quantization perturbs full-precision internal representations. For a given input, let \(H_l^{\mathrm{fp}}\) and \(H_l^{\mathrm{q}}\) denote the full-precision and quantized hidden states at layer \(l\), and let \(\mathcal{I}\) denote the set of valid token positions. For each channel \(c\), we define the channel-wise SNR as
\begin{equation}
    \mathrm{SNR}_{l,c}
    =
    10\log_{10}
    \frac{\sum_{p\in\mathcal{I}}\left(H_{l,p,c}^{\mathrm{fp}}\right)^2}
    {\sum_{p\in\mathcal{I}}\left(H_{l,p,c}^{\mathrm{q}}-H_{l,p,c}^{\mathrm{fp}}\right)^2},
    \label{eq:channel_snr}
\end{equation}
which measures the strength of the full-precision signal in that channel relative to the quantization error.

We also compute a global-energy SNR, which aggregates signal and error energies across all layers, token positions, and channels before taking their ratio:
\begin{equation}
    \mathrm{SNR}_{\mathrm{global}}
    =
    10\log_{10} \frac{\sum_l\sum_{p\in\mathcal{I}}\sum_c\left(H_{l,p,c}^{\mathrm{fp}}\right)^2}
    {\sum_l\sum_{p\in\mathcal{I}}\sum_c\left(H_{l,p,c}^{\mathrm{q}}-H_{l,p,c}^{\mathrm{fp}}\right)^2}.
    \label{eq:global_snr}
\end{equation}

We report SNR in two complementary forms. For channel-percentile summaries, we compute the 25th, 50th, 75th, and 99th percentiles of \(\mathrm{SNR}_{l,c}\) across channels within each layer, and then take the median across layers. These are reported as Median Ch25, Ch50, Ch75, and Ch99. The global-energy SNR aggregates signal and error energies across all layers and channels into a single energy ratio. We report both because representation errors are highly uneven across channels and layers: a single global ratio can be dominated by a few high-energy channels, whereas channel-percentile summaries avoid this influence. Input construction for the SNR analyses is detailed in Supplementary Table~\ref{tab:app_internal_inputs}.

\paragraph{Controlled pairwise comparison.}
We assess whether SNR aligns with performance retention through a controlled pairwise comparison. In each comparison, a single factor varies while all others are fixed under the same task. The three target factors are bit-width, quantization method, and model scale. For bit-width comparisons, the model and method are fixed. For method comparisons, the model and bit-width are fixed. For scale comparisons, models of different sizes within the same family are compared under the same method and bit-width. For a factor with $n$ distinct values, its $n$ settings are compared pairwise, yielding $n(n-1)/2$ candidate pairs before filtering.

For a pair of settings \(u\) and \(v\), it is counted as agreeing when the setting with higher retention also yields higher SNR, that is, when \(\mathrm{SNR}_u-\mathrm{SNR}_v\) and \(\mathrm{Ret}_u-\mathrm{Ret}_v\) have the same sign. Pairs with \(|\mathrm{Ret}_u-\mathrm{Ret}_v|<1.0\) percentage point are excluded as near ties. Let \(\mathcal{P}\) denote the set of valid pairs, and the agreement rate is defined as
\begin{equation}
    \mathrm{Agreement}
    =
    \frac{100}{|\mathcal{P}|}
    \sum_{(u,v)\in\mathcal{P}}
    \mathbf{1}\!\left[
    \left(\mathrm{SNR}_u-\mathrm{SNR}_v\right)
    \left(\mathrm{Ret}_u-\mathrm{Ret}_v\right)>0
    \right].
\end{equation}

If SNR and retention orderings are unrelated, each valid pair has an equal chance of agreeing or disagreeing, giving an expected agreement rate of \(50\%\). Let \(k\) be the number of agreeing pairs and \(n=|\mathcal{P}|\). We compute a one-sided binomial \(p\)-value under \(k\sim\mathrm{Binomial}(n,0.5)\), testing whether the observed agreement is significantly above 50\%. Stars in Fig.~\ref{fig:pairwise_comparison} denote significance levels (\(p<0.05\), \(p<0.01\) and \(p<0.001\)).

\subsection{Source SNR decomposition}
\label{sec:methods_msn}

\paragraph{Source SNR.}
For each linear module, let \(w_{l,c}\) denote the full-precision weight vector for output
channel \(c\) at layer \(l\), and let \(\delta w_{l,c}^{(a,b)}=\hat{w}_{l,c}^{(a,b)}-w_{l,c}\) be the perturbation introduced by quantization method \(a\) at bit-width \(b\). Given a full-precision input activation \(x_{l,p}\) at a valid token position \(p\), the full-precision output is \(w_{l,c}^{\top}x_{l,p}\), and the quantization error is \(\delta w_{l,c}^{(a,b)\top}x_{l,p}\). For this token, the signal energy is therefore \((w_{l,c}^{\top}x_{l,p})^2\), and the error energy is \((\delta w_{l,c}^{(a,b)\top}x_{l,p})^2\). Aggregating these energies over the valid token positions of task \(t\) using the trace‑normalized activation second moment \(\bar{\Sigma}_{x,l}^{(t)}\) gives the normalized signal energy \(w_{l,c}^{\top}\bar{\Sigma}_{x,l}^{(t)}w_{l,c}\) and the normalized error energy \(\delta w_{l,c}^{(a,b)\top}\bar{\Sigma}_{x,l}^{(t)}\delta w_{l,c}^{(a,b)}\). The source SNR in decibels is therefore
\begin{equation}
    \mathrm{SNR}_{l,c}^{\mathrm{source}}
    =
    10\log_{10}
    \frac{w_{l,c}^{\top}\bar{\Sigma}_{x,l}^{(t)}w_{l,c}}
    {\delta w_{l,c}^{(a,b)\top}\bar{\Sigma}_{x,l}^{(t)}\delta w_{l,c}^{(a,b)}} .
\end{equation}
Trace normalization ensures a consistent activation scale across tasks without affecting the ratio, as the same factor appears in both numerator and denominator. On this log scale, the ratio decomposes additively as
\begin{equation}
    \mathrm{SNR}_{l,c}^{\mathrm{source}} = M_{l,c}+S_{l,c}-N_{l,c},
\end{equation}
where
\begin{equation}
    \begin{aligned}
        M_{l,c} &= 10\log_{10}\frac{\|w_{l,c}\|_2^2}{\|\delta w_{l,c}^{(a,b)}\|_2^2},\\
        S_{l,c} &= 10\log_{10}\frac{w_{l,c}^{\top}\bar{\Sigma}_{x,l}^{(t)}w_{l,c}}{\|w_{l,c}\|_2^2},\\
        N_{l,c} &= 10\log_{10}\frac{\delta w_{l,c}^{(a,b)\top}\bar{\Sigma}_{x,l}^{(t)}\delta w_{l,c}^{(a,b)}}{\|\delta w_{l,c}^{(a,b)}\|_2^2}.
    \end{aligned}
\end{equation}
\(M\) measures the magnitude of the original weight relative to its quantization error, with larger values indicating a smaller perturbation. \(S\) measures the energy of the task-specific activations along the direction of the full-precision weight, whereas \(N\) measures the activation energy along the direction of the quantization error. Larger \(M\) and \(S\) increase the source SNR, whereas larger \(N\) decreases it.


\paragraph{Modules and aggregation.}
The decomposition is applied to all seven linear modules in each transformer block: the query, key, value and output projections in attention, and the gate, up and down projections in the MLP. For each module, \(M\), \(S\), \(N\) and source SNR are computed per channel. In Fig.~\ref{fig:msn}, we take the channel median at each layer and module, followed by the median across layers and then across modules. This stepwise median aggregation reduces the influence of outlier layers and modules. Results for individual modules and sensitivity to alternative channel and layer summaries are provided in Supplementary Fig.~\ref{fig:app_modulewise_source_snr} and Supplementary Table~\ref{tab:app_summary_statistic_robustness}, respectively.

\subsection{Source--propagation analysis of accumulated NSR}
\label{sec:methods_source_propagation}

We analyze cross-layer error accumulation using the noise-to-signal ratio (NSR), the inverse energy form of SNR. Whereas the source SNR decomposition in Section~\ref{sec:methods_msn} measures errors introduced by quantized modules under clean inputs, the accumulated NSR analysis separates the error propagated from earlier layers from the error newly introduced by quantizing the current layer.

\paragraph{Error decomposition.}
For each transformer block \(l\), let \(H_l^{\mathrm{fp}}\) and \(H_l^{\mathrm{q}}\) denote the full-precision and quantized hidden states entering the block, and let \(F_l^{\mathrm{fp}}\) and \(F_l^{\mathrm{q}}\) denote the corresponding block functions. The resulting error is
\begin{equation}
    T_l = F_l^{\mathrm{q}}(H_l^{\mathrm{q}}) - F_l^{\mathrm{fp}}(H_l^{\mathrm{fp}}).
\end{equation}

This error arises from two distinct parts: the new error introduced by quantizing block \(l\) itself, and the error already accumulated in earlier layers, which may be further altered as it passes through the block. To separate these contributions, we compare two controlled scenarios: passing the clean representation through the quantized block, which captures the new error introduced by quantization, and passing the quantized representation through the full-precision block, which captures how the existing error evolves. This yields an exact decomposition
\begin{equation}
    \begin{aligned}
        T_l
        &=
        \underbrace{\left[
        F_l^{\mathrm{fp}}(H_l^{\mathrm{q}})
        - F_l^{\mathrm{fp}}(H_l^{\mathrm{fp}})
        \right]}_{\text{propagated error } P_l}
        +
        \underbrace{\left[
        F_l^{\mathrm{q}}(H_l^{\mathrm{fp}})
        - F_l^{\mathrm{fp}}(H_l^{\mathrm{fp}})
        \right]}_{\text{source error } C_l} \\
        &\quad+
        \underbrace{\left[
        F_l^{\mathrm{q}}(H_l^{\mathrm{q}})
        - F_l^{\mathrm{fp}}(H_l^{\mathrm{q}})
        - F_l^{\mathrm{q}}(H_l^{\mathrm{fp}})
        + F_l^{\mathrm{fp}}(H_l^{\mathrm{fp}})
        \right]}_{\text{interaction error } I_l}.
    \end{aligned}
\end{equation}
Here, \(P_l\) is the accumulated error propagated through the full-precision block, \(C_l\) is the source error introduced by quantizing the current block \(l\) under clean activations, and \(I_l\) is the residual interaction not explained by these two components alone.


\paragraph{NSR and propagation gain.}
The decomposition above separates the output error into propagated, source and interaction components. These components are converted to NSR form by normalizing each squared error by the corresponding full-precision signal energy. The accumulated NSR before and after block \(l\) are
\begin{equation}
    R_l = \frac{\|H_l^{\mathrm{q}}-H_l^{\mathrm{fp}}\|_F^2}
    {\|H_l^{\mathrm{fp}}\|_F^2}, \qquad R_{l+1} = \frac{\|T_l\|_F^2}
    {\|F_l^{\mathrm{fp}}(H_l^{\mathrm{fp}})\|_F^2}.
\end{equation}

The resulting NSR components are
\begin{equation}
    R_l^{\mathrm{prop}} =
    \frac{\|P_l\|_F^2}{\|F_l^{\mathrm{fp}}(H_l^{\mathrm{fp}})\|_F^2},
    \quad
    Q_l =
    \frac{\|C_l\|_F^2}{\|F_l^{\mathrm{fp}}(H_l^{\mathrm{fp}})\|_F^2},
    \quad
    R_l^I =
    \frac{\|I_l\|_F^2}{\|F_l^{\mathrm{fp}}(H_l^{\mathrm{fp}})\|_F^2}.
\end{equation}
Here, \(Q_l\) denotes the source NSR introduced by the current block.

For each block after the first (\(l=1,\ldots,L-1\)), the propagation gain is defined as
\begin{equation}
    G_l = \frac{R_l^{\mathrm{prop}}}{R_l} = \frac{\|F_l^{\mathrm{fp}}(H_l^{\mathrm{q}})-F_l^{\mathrm{fp}}(H_l^{\mathrm{fp}})\|_F^2 / \|F_l^{\mathrm{fp}}(H_l^{\mathrm{fp}})\|_F^2}{\|H_l^{\mathrm{q}}-H_l^{\mathrm{fp}}\|_F^2 / \|H_l^{\mathrm{fp}}\|_F^2}.
\end{equation}

It measures how the accumulated error is altered after passing through block \(l\), with \(G_l>1\) corresponding to error amplification, \(G_l<1\) to error attenuation, and \(G_l\approx1\) meaning that the error passes through largely unchanged.

\paragraph{Layer-wise recurrence.}
Starting from the exact error decomposition \(T_l=P_l+C_l+I_l\), we expand the squared norm as
\begin{equation}
    \|T_l\|_F^2
    =
    \|P_l\|_F^2+\|C_l\|_F^2+\|I_l\|_F^2
    +2\langle P_l,C_l\rangle
    +2\langle P_l,I_l\rangle
    +2\langle C_l,I_l\rangle .
\end{equation}
Dividing both sides by the output energy of the full-precision model
\(\|F_l^{\mathrm{fp}}(H_l^{\mathrm{fp}})\|_F^2\), and using
\(R_l^{\mathrm{prop}}=G_lR_l\), gives
\begin{equation}
    R_{l+1}=G_lR_l+Q_l+D_l,
\end{equation}
where
\begin{equation}
    D_l = \frac{\|I_l\|_F^2 + 2\langle P_l,C_l\rangle + 2\langle P_l,I_l\rangle + 2\langle C_l,I_l\rangle}{\|F_l^{\mathrm{fp}}(H_l^{\mathrm{fp}})\|_F^2}.
\end{equation}
The term \(D_l\) contains the residual contributions beyond \(G_lR_l\) and \(Q_l\): the energy of the interaction term \(I_l\) and the pairwise overlaps among \(P_l\), \(C_l\) and \(I_l\).

To examine how much of the accumulated NSR can be accounted for by the propagated and source components alone, we use a reduced recurrence that omits \(D_l\):
\begin{equation}
    \widehat{R}_{l+1}=G_l\widehat{R}_l+Q_l,
    \qquad
    \widehat{R}_0=0 .
\end{equation}
The reduced recurrence starts from \(\widehat{R}_0=0\) and is iterated over \(L\) layers indexed by \(l=0,\ldots,L-1\). Unrolling the recurrence to the output of the final layer gives
\begin{equation}
    \widehat{R}_L = \sum_{l=0}^{L-1} \left( \prod_{j=l+1}^{L-1}G_j \right)Q_l .
\end{equation}
This expression shows that the source NSR \(Q_l\) introduced at layer \(l\) contributes to the final accumulated NSR after being scaled by the propagation gains of subsequent layers. We compare \(\widehat{R}_l\) with \(R_l\) across depth to evaluate how well this reduced recurrence reconstructs the observed accumulated NSR.

\paragraph{Aggregation.}
Panels~\textbf{a}--\textbf{c} in Fig.~\ref{fig:source_propagation} use global Frobenius-energy ratios, obtained by summing squared errors over all valid tokens and representation dimensions to form the NSR terms. For panel~\textbf{a}, we first compute the relative shares of \(R_l^{\mathrm{prop}}\), \(Q_l\), and \(R_l^I\) for each layer. These layer-wise shares are then averaged within each depth stage (early, middle, and late thirds of normalized depth) and re-normalized to sum to one. We use \(R_l^I\) rather than \(D_l\) because \(D_l\) can be negative due to signed overlap terms, making it unsuitable for computing relative shares. Panels~\textbf{d} and~\textbf{e} report channel-median NSR, computed for each layer by taking the median across channels.


\section*{Data availability}
The benchmark datasets used in this study are publicly available from their original sources.

\section*{Code availability}
The COSQuant toolkit developed in this study is publicly available at \url{https://github.com/Cfish808/LLM-Compression-Tool}. The code used for the source SNR decomposition and cross-layer propagation analyses will be made publicly available upon publication.

\bibliographystyle{conference-numeric}
\bibliography{conference}

\input{supplementary_information}

\end{document}

%% file: supplementary_information.tex
\clearpage
\appendix
\setcounter{figure}{0}
\setcounter{table}{0}
\setcounter{equation}{0}
\renewcommand{\thefigure}{\arabic{figure}}
\renewcommand{\thetable}{\arabic{table}}
\renewcommand{\theequation}{S\arabic{equation}}
\renewcommand{\theHfigure}{supp.\arabic{figure}}
\renewcommand{\theHtable}{supp.\arabic{table}}
\renewcommand{\theHequation}{supp.S\arabic{equation}}
\captionsetup[figure]{name=Supplementary Fig.}
\captionsetup[table]{name=Supplementary Table}

\section{Experimental details}
\label{app:experimental_setup}

\subsection{Models and quantization settings}
\label{app:model_quant_coverage}

Supplementary Table~\ref{tab:app_model_coverage} summarizes the model families and scales used in our study. Llama-2, Llama-3, Mistral, and Qwen2.5 are used for consistency evaluation across families. The Qwen3 series offers five model scales spanning 0.6B to 14B, with its 8B and 14B models also contributing to the cross-family consistency evaluation.

\begin{table}[h]
    \centering
    \caption{
        \textbf{Model coverage.}
        Model families and scales used for the degradation analyses.
    }
    \label{tab:app_model_coverage}
    \begin{tabular}{lll}
        \toprule
        Family & Model scales & Role in analysis \\
        \midrule
        Llama-2 & Llama-2-7B, Llama-2-13B & Consistency evaluation across families \\
        Llama-3 & Llama-3-8B & Consistency evaluation across families \\
        Mistral & Mistral-7B & Consistency evaluation across families \\
        Qwen2.5 & Qwen2.5-7B & Consistency evaluation across families \\
        Qwen3 & Qwen3-0.6B, 1.7B, 4B, 8B, 14B & Model-scale analysis and cross-family evaluation \\
        \bottomrule
    \end{tabular}
\end{table}

Supplementary Table~\ref{tab:app_quant_methods} summarizes the quantization methods, bit-widths, model coverage and roles in analysis. GPTQ is evaluated on all ten models, whereas RTN, AWQ and SPQR are each evaluated on the same six models: Llama-2-7B, Llama-2-13B, Llama-3-8B, Qwen2.5-7B, Qwen3-8B and Qwen3-14B. Together, these settings yield 76 quantized configurations: 40 from GPTQ (10 models \(\times\) 4 bit-widths) and 36 from RTN, AWQ and SPQR (6 models \(\times\) 3 methods \(\times\) 2 bit-widths).

\begin{table}[h]
    \centering
    \caption{
        \textbf{Quantization settings.}
        Bit-widths and model coverage used for each weight-only PTQ method.
    }
    \label{tab:app_quant_methods}
    \begin{tabular}{llll}
        \toprule
        Method & Bit-widths & Model coverage & Role in analysis \\
        \midrule
        GPTQ & W8, W4, W3, W2 & All ten models &
        Bit-width comparison and model-scale analysis \\
        RTN & W4, W3 & Six models & Method comparison \\
        AWQ & W4, W3 & Six models & Method comparison \\
        SPQR & W4, W3 & Six models & Method comparison \\
        \bottomrule
    \end{tabular}
\end{table}

\subsection{Downstream benchmarks and evaluation protocol}
\label{app:benchmarks_retention}

\begin{table}[!t]
    \centering
    \caption{
        \textbf{Downstream benchmarks for the evaluation.}
        MC denotes multiple-choice evaluation and Gen denotes open-ended generation.
    }
    \label{tab:app_downstream_benchmarks}
    \begin{tabular}{ccccc}
        \toprule
        Benchmark & Task type & Evaluation format & Metric & Random baseline \\
        \midrule
        WinoGrande & Commonsense reasoning & MC (2) & Acc & 0.50 \\
        HellaSwag & Commonsense reasoning & MC (4) & Acc & 0.25 \\
        ARC-Easy & Commonsense reasoning & MC (4) & Acc & 0.25 \\
        PIQA & Commonsense reasoning & MC (2) & Acc & 0.50 \\
        SocialIQA & Commonsense reasoning & MC (3) & Acc & 0.33 \\
        \midrule
        TriviaQA & Knowledge-intensive & Gen & EM & \(\approx 0\) \\
        MMLU & Knowledge-intensive & MC (4) & Acc & 0.25 \\
        \midrule
        ARC-Challenge & Reasoning & MC (4) & Acc & 0.25 \\
        StrategyQA & Reasoning & MC (2) & Acc & 0.50 \\
        MathQA & Reasoning & MC (5) & Acc & 0.20 \\
        GSM8K & Reasoning & Gen & EM & \(\approx 0\) \\
        \bottomrule
    \end{tabular}
\end{table}

Supplementary Table~\ref{tab:app_downstream_benchmarks} lists the 11 downstream benchmarks used for the cross-family and quantization-method comparisons, together with their task types, evaluation formats and random baselines for computing de-randomized retention. The Qwen3 model-scale analysis uses nine benchmarks across all five model scales: WinoGrande, HellaSwag and ARC-Easy for commonsense reasoning; TriviaQA and MMLU for knowledge-intensive tasks; and StrategyQA, GSM8K, MathQA and ARC-Challenge for reasoning tasks. For both evaluation sets, retention scores are first averaged within each task type. These scores are then averaged across task types with equal weight to obtain the overall downstream score, preventing task groups with more benchmarks from dominating the result.

\subsection{Inputs for signal-to-noise analyses}
\label{app:internal_inputs}

The internal signal-to-noise analyses use task-specific inputs constructed from WikiText-2, HellaSwag and GSM8K. For each analysis, the full-precision and quantized models process the same tokenized inputs under an identical attention mask. Signal and error energies are accumulated over non-padding tokens only.

Table~\ref{tab:app_internal_inputs} summarizes the input construction. Each task contributes 128 examples, with a maximum sequence length of 1024 and a fixed random seed of 42. WikiText-2 uses contiguous token blocks from the test split and requires no random sampling. HellaSwag and GSM8K are shuffled with the fixed seed before selecting examples. For HellaSwag, each context is paired with four candidate endings; the model processes each context–ending pair as an independent sequence, producing four sequences per example. GSM8K uses a fixed five-shot chain-of-thought prefix, followed by the test question and an answer cue.

\begin{table}[!t]
    \centering
    \caption{
        \textbf{Task-specific input construction for internal signal-to-noise analyses.} WikiText-2, HellaSwag and GSM8K represent language modelling, commonsense reasoning and mathematical reasoning, respectively.
    }
    \label{tab:app_internal_inputs}
    \begin{tabular}{@{}lllc@{}}
        \toprule
        Task & Split & Input sequence & Examples/sequences \\
        \midrule
        WikiText-2 & Test & Contiguous 1024-token blocks from the concatenated test text & 128 / 128 \\
        HellaSwag & Validation & Context concatenated with each candidate ending & 128 / 512 \\
        GSM8K & Test & Fixed five-shot CoT prefix, test question and answer cue & 128 / 128 \\
        \bottomrule
    \end{tabular}
\end{table}

\section{Detailed evaluation results}
\label{app:detailed_evaluation_results}

The main text compares quantization degradation across bit-widths, quantization methods and model scales using retention averaged within commonsense reasoning, knowledge-intensive and reasoning tasks. Here, we report results for each benchmark and show how the Qwen3 scaling trend differs among the three task types.

\subsection{Retention across individual downstream tasks}
\label{app:task_level_degradation}

\begin{figure}[p]
    \centering
    \includegraphics[width=\textwidth,height=0.91\textheight,keepaspectratio]{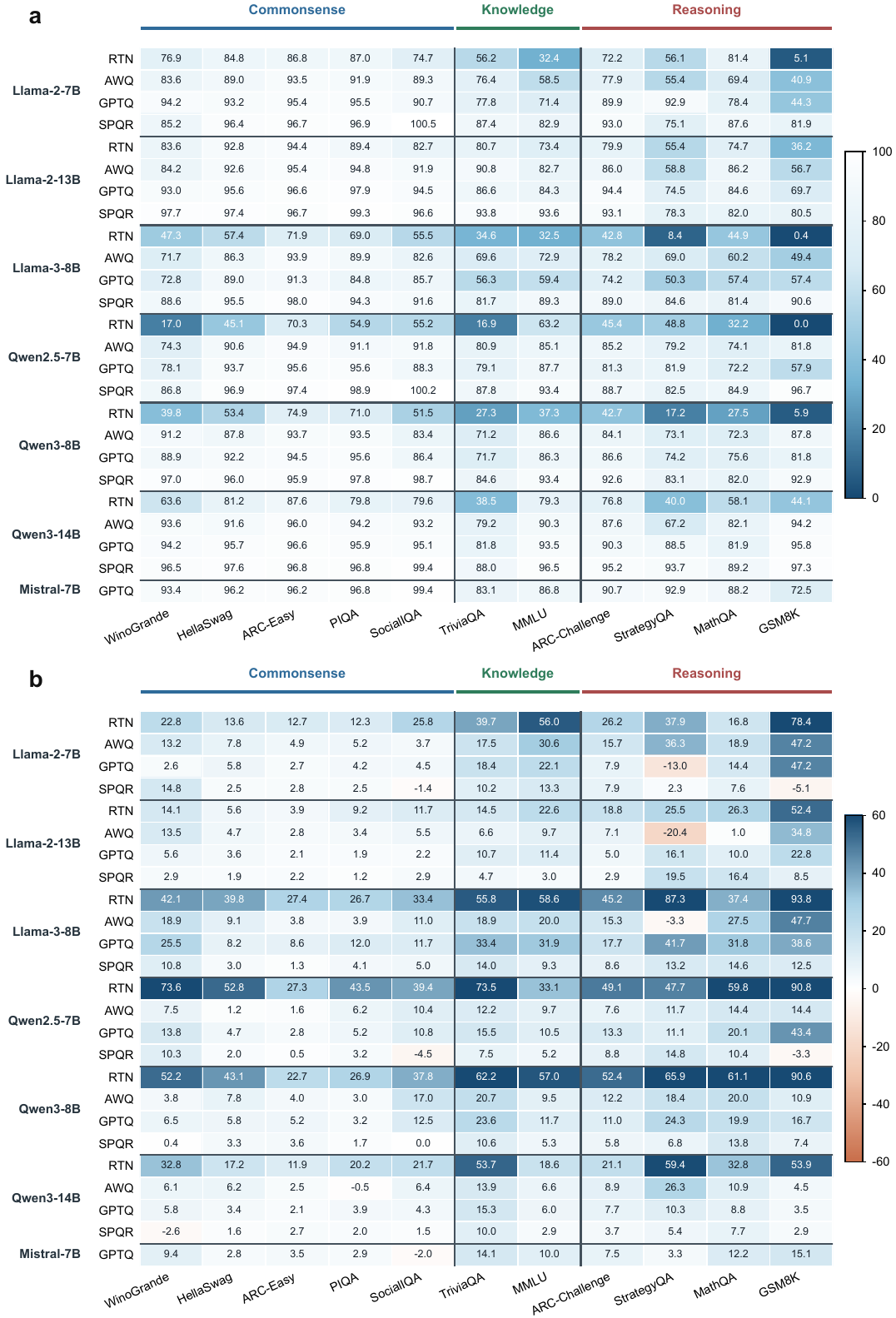}
    \caption{
        \textbf{W3 retention and W4-to-W3 retention decrease across downstream benchmarks.}
        Rows correspond to the models and quantization methods used, while columns show the 11 downstream benchmarks grouped into three task types.
        \textbf{a}, De-randomized W3 retention relative to the corresponding full-precision model.
        \textbf{b}, W4-to-W3 retention decrease, calculated as W4 retention minus W3 retention in percentage points. Darker blue denotes lower retention in \textbf{a} and a larger decrease in \textbf{b}; orange cells in \textbf{b} indicate occasional higher retention at W3 than at W4.
    }
    \label{fig:app_task_level_retention}
\end{figure}

Supplementary Fig.~\ref{fig:app_task_level_retention} presents W3 retention and the W4-to-W3 retention decrease for each of the 11 downstream benchmarks. The results support the aggregate trends in main-text Fig.~1a,b: at W3, performance is better preserved on commonsense reasoning tasks than on knowledge-intensive and reasoning tasks. RTN produces substantial W4-to-W3 decreases across the widest range of benchmarks, whereas AWQ, GPTQ and SPQR generally retain more performance, with their relative performance varying across models and tasks.

\clearpage

\subsection{Scale-dependent retention across task types}
\label{app:scale_degradation_by_task_type}

\begin{figure}[!t]
    \centering
    \includegraphics[width=\textwidth]{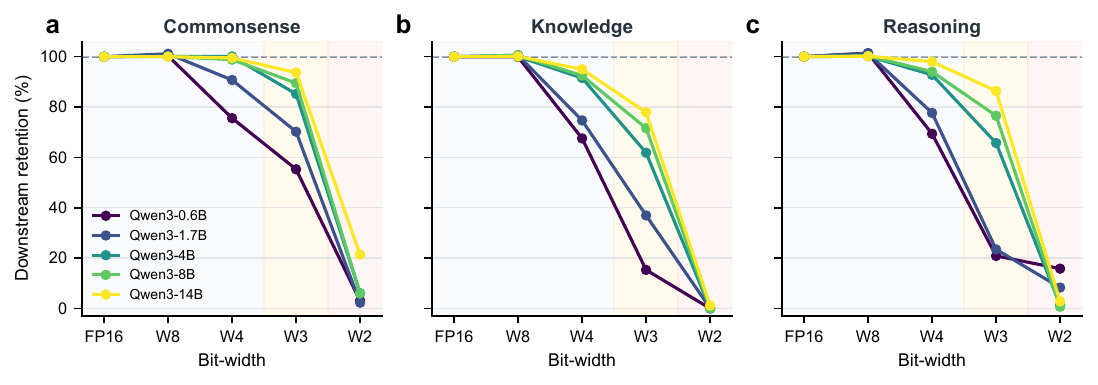}
    \caption{
        \textbf{Scale-dependent downstream retention across task types.}
        Qwen3 models from 0.6B to 14B are quantized with GPTQ and evaluated at W8, W4, W3 and W2.
        \textbf{a}, Commonsense retention, averaged over WinoGrande, HellaSwag and ARC-Easy.
        \textbf{b}, Knowledge retention, averaged over TriviaQA and MMLU.
        \textbf{c}, Reasoning retention, averaged over StrategyQA, GSM8K, MathQA and ARC-Challenge.
        De-randomized retention is computed separately for each benchmark and then averaged within each task type. The dashed horizontal line marks the full-precision reference at 100\%.
    }
    \label{fig:app_scale_retention_by_task_type}
\end{figure}

Supplementary Fig.~\ref{fig:app_scale_retention_by_task_type} presents the Qwen3 scaling result in main-text Fig.~1c separately for the three task types. Retention generally increases with model scale at W4 and W3, with differences between scales growing markedly at W3, especially for knowledge-intensive and reasoning tasks. At W2, all models and task types degrade sharply. The higher overall retention of larger models is thus present across all task types at W3 but most pronounced for knowledge-intensive and reasoning tasks.

\clearpage

\section{Source SNR analysis}
\label{app:source_snr_decomposition}

This section provides the derivation and implementation details of the source SNR decomposition, examines its behavior across quantized linear modules, and evaluates its sensitivity to the summary statistics used in the main analysis.

\subsection{Derivation and implementation}
\label{app:source_snr_derivation}

We present the derivation for one output channel of a quantized linear module; the same computation is applied independently to every output channel. Consider a linear module in transformer block \(l\). Let \(w_{l,c}\) be its full-precision weight vector for output channel \(c\), and let \(\delta w_{l,c}^{(a,b)}=\hat{w}_{l,c}^{(a,b)}-w_{l,c}\) be the weight quantization error introduced by method \(a\) at bit-width \(b\). For a full-precision input activation \(x_{l,p}\) at valid token position \(p\), the clean scalar response and newly introduced source error are
\begin{equation}
    y_{l,p,c}=w_{l,c}^{\top}x_{l,p},
    \qquad
    e_{l,p,c}=\delta w_{l,c}^{(a,b)\top}x_{l,p}.
\end{equation}

To aggregate these per-token responses across all valid tokens of task \(t\), we define the empirical second moment of the full-precision activations as
\begin{equation}
    \Sigma_{x,l}^{(t)}
    =
    \frac{1}{|\mathcal{I}^{(t)}|}
    \sum_{p\in\mathcal{I}^{(t)}}x_{l,p}x_{l,p}^{\top},
\end{equation}
where \(\mathcal{I}^{(t)}\) denotes the set of valid non-padding token positions for task \(t\). The channel-level source energy ratio is then
\begin{equation}
    \frac{\mathbb{E}_{p\in\mathcal{I}^{(t)}}[y_{l,p,c}^{2}]}
    {\mathbb{E}_{p\in\mathcal{I}^{(t)}}[e_{l,p,c}^{2}]}
    =
    \frac{w_{l,c}^{\top}\Sigma_{x,l}^{(t)}w_{l,c}}
    {\delta w_{l,c}^{(a,b)\top}\Sigma_{x,l}^{(t)}\delta w_{l,c}^{(a,b)}}.
\end{equation}
This equality is exact and places no assumptions on the distribution of the quantization error.

To remove the overall activation-energy scale while preserving the source energy ratio, we use the trace-normalized second moment
\begin{equation}
    \bar{\Sigma}_{x,l}^{(t)}
    =
    \frac{\Sigma_{x,l}^{(t)}}{\operatorname{Tr}\!\left(\Sigma_{x,l}^{(t)}\right)}.
\end{equation}
Because the same trace factor appears in both the numerator and the denominator, replacing \(\Sigma_{x,l}^{(t)}\) with \(\bar{\Sigma}_{x,l}^{(t)}\) leaves the source energy ratio unchanged. Multiplying and dividing by the squared norms of \(w_{l,c}\) and \(\delta w_{l,c}^{(a,b)}\) gives
\begin{equation}
    \frac{w_{l,c}^{\top}\bar{\Sigma}_{x,l}^{(t)}w_{l,c}}
    {\delta w_{l,c}^{(a,b)\top}\bar{\Sigma}_{x,l}^{(t)}\delta w_{l,c}^{(a,b)}}
    =
    \frac{\|w_{l,c}\|_2^2}{\|\delta w_{l,c}^{(a,b)}\|_2^2}
    \cdot
    \frac{
    w_{l,c}^{\top}\bar{\Sigma}_{x,l}^{(t)}w_{l,c}/\|w_{l,c}\|_2^2}
    {
    \delta w_{l,c}^{(a,b)\top}\bar{\Sigma}_{x,l}^{(t)}
    \delta w_{l,c}^{(a,b)}/\|\delta w_{l,c}^{(a,b)}\|_2^2}.
\end{equation}

Taking \(10\log_{10}(\cdot)\) yields the additive source SNR decomposition used in the main text,
\begin{equation}
    \mathrm{SNR}_{l,c}^{\mathrm{source}}
    =
    M_{l,c}+S_{l,c}-N_{l,c},
\end{equation}
where
\begin{equation}
    \begin{aligned}
        M_{l,c}
        &=
        10\log_{10}
        \frac{\|w_{l,c}\|_2^2}
        {\|\delta w_{l,c}^{(a,b)}\|_2^2}, \\
        S_{l,c}
        &=
        10\log_{10}
        \frac{w_{l,c}^{\top}\bar{\Sigma}_{x,l}^{(t)}w_{l,c}}
        {\|w_{l,c}\|_2^2}, \\
        N_{l,c}
        &=
        10\log_{10}
        \frac{\delta w_{l,c}^{(a,b)\top}\bar{\Sigma}_{x,l}^{(t)}\delta w_{l,c}^{(a,b)}}
        {\|\delta w_{l,c}^{(a,b)}\|_2^2}.
    \end{aligned}
\end{equation}

\(M\) measures the magnitude of the original weight relative to its quantization error, with larger values indicating a smaller relative error. The signal term \(S\) measures the energy of the task-specific activations along the direction of the full-precision weight. The noise term \(N\) measures the activation energy along the direction of the quantization error. Larger \(M\) and \(S\) increase the source SNR, whereas larger \(N\) decreases it.

\paragraph{Implementation and aggregation.}
The implementation does not explicitly construct \(\Sigma_{x,l}^{(t)}\). Instead, it iterates over the valid token positions and accumulates the squared clean response \((w_{l,c}^{\top}x_{l,p})^2\), the squared quantization-error response \((\delta w_{l,c}^{(a,b)\top}x_{l,p})^2\), and the input energy \(\|x_{l,p}\|_2^2\). The normalized signal and error energies can therefore be computed directly as the ratios of the accumulated response energies to the accumulated input energy:
\begin{equation}
    \begin{aligned}
        w_{l,c}^{\top}\bar{\Sigma}_{x,l}^{(t)}w_{l,c}
        &=
        \frac{
        \sum_{p\in\mathcal{I}^{(t)}}(w_{l,c}^{\top}x_{l,p})^2}
        {\sum_{p\in\mathcal{I}^{(t)}}\|x_{l,p}\|_2^2},
        \\
        \delta w_{l,c}^{(a,b)\top}\bar{\Sigma}_{x,l}^{(t)}
        \delta w_{l,c}^{(a,b)}
        &=
        \frac{
        \sum_{p\in\mathcal{I}^{(t)}}
        (\delta w_{l,c}^{(a,b)\top}x_{l,p})^2}
        {\sum_{p\in\mathcal{I}^{(t)}}\|x_{l,p}\|_2^2}.
    \end{aligned}
\end{equation}
This computation produces the same normalized energies as the matrix formulation without storing a \(d_{\mathrm{in}}\times d_{\mathrm{in}}\) activation second-moment matrix.

The decomposition covers all seven quantized linear modules in each transformer block: the query, key, value and output projections in self-attention, and the gate, up and down projections in the MLP. At each layer and module, \(M\), \(S\), \(N\) and source SNR are computed for every output channel. The main analysis then takes the median across channels, the median across layers within each module, and finally the median across modules. We also compute the channel percentiles Ch25, Ch75 and Ch99 for the sensitivity analysis in Supplementary Section~\ref{app:source_snr_sensitivity}.

\subsection{Module-wise source SNR components}
\label{app:modulewise_source_snr}

\begin{figure}[p]
    \centering
    \includegraphics[width=\textwidth,height=0.86\textheight,keepaspectratio]{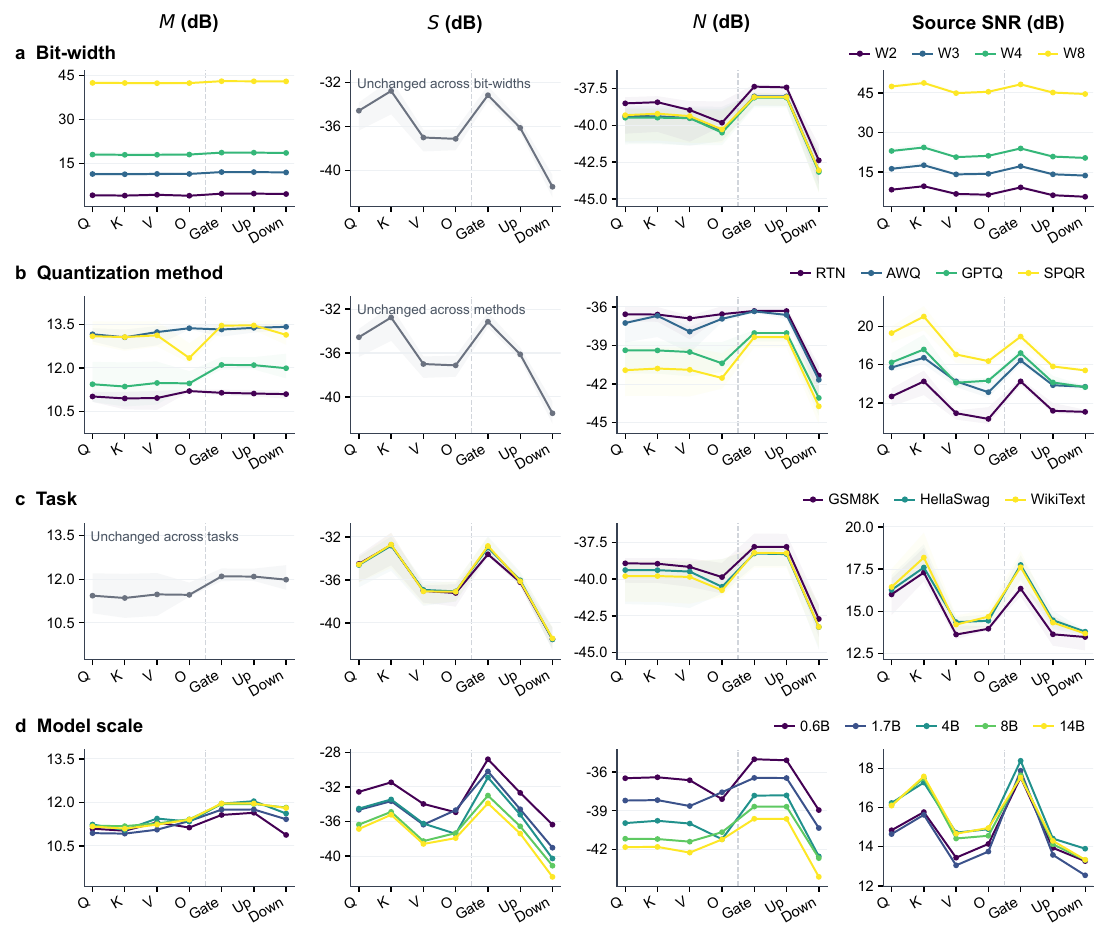}
    \caption{
        \textbf{Module-wise source SNR decomposition across quantization settings, tasks and model scales.}
        Columns report \(M\), \(S\), \(N\) and source SNR for the query (Q), key (K), value (V), output (O), gate, up and down projections.
        \textbf{a}, GPTQ bit-width comparison.
        \textbf{b}, Quantization-method comparison at W3.
        \textbf{c}, Task comparison under GPTQ-W3.
        \textbf{d}, Qwen3 model-scale comparison under GPTQ-W3.
        Panels \textbf{a}--\textbf{c} use Llama-2-7B, Llama-3-8B, Qwen2.5-7B and Qwen3-8B. Values are channel medians followed by medians across layers; panels \textbf{a} and \textbf{b} are additionally averaged across WikiText-2, HellaSwag and GSM8K. Lines denote medians across the four models, and shading denotes their range. Panel \textbf{d} reports each Qwen3 model after averaging across the same three datasets. When a quantity is unchanged across the compared conditions, the overlapping curves are shown as a single grey line. Lines between modules are included only as visual guides.
    }
    \label{fig:app_modulewise_source_snr}
\end{figure}

The main analysis aggregates the source SNR components across the seven quantized linear modules. Supplementary Fig.~\ref{fig:app_modulewise_source_snr} instead reports \(M\), \(S\), \(N\) and source SNR separately for the query, key, value, output, gate, up and down projections, allowing direct comparison of how bit-width, quantization method, task and model scale affect each module.

Reducing bit-width lowers \(M\) in every module, while \(S\) is unchanged and \(N\) changes only slightly. At W3, quantization methods differ in both \(M\) and \(N\) across the seven modules. Changing the task leaves \(M\) unchanged but alters \(S\) and \(N\), while model scale affects all three components. Although the absolute values differ among projections, the main-text conclusions are not determined by a single module, supporting the use of the module median.

\clearpage

\subsection{Sensitivity to summary statistics}
\label{app:source_snr_sensitivity}

The main analysis summarizes \(M\), \(S\), \(N\) and source SNR using the channel median (Ch50), followed by medians across layers and modules. To assess sensitivity to these aggregation choices, we repeat the analysis under four alternative summaries: replacing Ch50 with Ch25, Ch75 or Ch99, or replacing the layer median with the layer mean, leaving the other aggregation steps unchanged.

The sensitivity analysis covers 67 quantized configurations across ten models (the Qwen3 models below 8B are evaluated only under GPTQ-W3). Each configuration is analyzed on WikiText-2, HellaSwag and GSM8K, yielding 201 configuration–task combinations. For each of \(M\), \(S\), \(N\) and source SNR, we compute the main summary and every alternative summary for all combinations. \(M\) does not vary across tasks for a given quantized configuration, so only one value per configuration is used, leaving 67 values. \(S\) does not vary across quantization methods or bit-widths for a given model--task pair, so only one value per pair is used, leaving 30 values. \(N\) and source SNR vary with all factors, so all 201 values are used. Spearman's \(\rho\) is then computed between the main and alternative summaries using the appropriate set of values for each quantity. A high correlation indicates that the alternative procedure largely preserves the ordering obtained with the main aggregation.

Supplementary Table~\ref{tab:app_summary_statistic_robustness} reports the correlations for each alternative. The rankings of \(M\), \(S\), \(N\) and source SNR remain stable across tested alternatives. The only notable deviation occurs for \(S\) under Ch99 (\(\rho=0.689\)), indicating that the ordering of \(S\) in the extreme upper tail can differ from the median-based ordering. Nevertheless, the conclusions based on source SNR remain robust, with correlations of at least 0.962 under every alternative.

\begin{table}[!t]
    \centering
    \caption{
        \textbf{Sensitivity of the source SNR analysis to alternative summary statistics.}
        Spearman rank correlations compare each alternative aggregation with the main procedure. Correlations are computed over 67 quantized configurations for \(M\), 30 model--task combinations for \(S\), and 201 configuration--task combinations for \(N\) and source SNR.
    }
    \label{tab:app_summary_statistic_robustness}
    \begin{tabular}{@{}lcccc@{}}
        \toprule
        Alternative & \(M\) & \(S\) & \(N\) & Source SNR \\
        \midrule
        Ch25 & 0.999 & 0.956 & 0.986 & 0.998 \\
        Ch75 & 0.998 & 0.961 & 0.988 & 0.995 \\
        Ch99 & 0.976 & 0.689 & 0.910 & 0.962 \\
        Layer mean & 0.998 & 0.957 & 0.985 & 0.995 \\
        \bottomrule
    \end{tabular}
\end{table}

\section{Source--propagation Analysis}
\label{app:source_propagation}

The source--propagation analysis separates error newly introduced by each transformer block from error accumulated in preceding blocks. The main text reports this analysis for WikiText-2; here, we provide the exact recurrence and evaluate the same quantities using HellaSwag and GSM8K.

\subsection{Error decomposition and recurrence}
\label{app:source_propagation_recurrence}

\paragraph{Block-level error decomposition.}
For transformer block \(l\), let \(H_l^{\mathrm{fp}}\) and \(H_l^{\mathrm{q}}\) denote the full-precision and quantized hidden states entering the block, and let \(F_l^{\mathrm{fp}}\) and \(F_l^{\mathrm{q}}\) denote the corresponding block functions. The output error is
\begin{equation}
    T_l=F_l^{\mathrm{q}}(H_l^{\mathrm{q}})-F_l^{\mathrm{fp}}(H_l^{\mathrm{fp}}),
\end{equation}
which can be decomposed exactly as
\begin{equation}
    T_l=P_l+C_l+I_l,
\end{equation}
where
\begin{equation}
    \begin{aligned}
        P_l &= F_l^{\mathrm{fp}}(H_l^{\mathrm{q}})-F_l^{\mathrm{fp}}(H_l^{\mathrm{fp}}),\\
        C_l &= F_l^{\mathrm{q}}(H_l^{\mathrm{fp}})-F_l^{\mathrm{fp}}(H_l^{\mathrm{fp}}),\\
        I_l &= F_l^{\mathrm{q}}(H_l^{\mathrm{q}})-F_l^{\mathrm{fp}}(H_l^{\mathrm{q}})-F_l^{\mathrm{q}}(H_l^{\mathrm{fp}})+F_l^{\mathrm{fp}}(H_l^{\mathrm{fp}}).
    \end{aligned}
\end{equation}
The propagated error \(P_l\) isolates the effect of the accumulated error at the block input, with the current block kept in full precision. The source error \(C_l\) isolates the effect of quantizing the current block when both block functions receive the same full-precision input. The interaction term \(I_l\) accounts for their non-additive effect when the accumulated input error and current-block quantization are present simultaneously.

\paragraph{NSR and propagation gain.}
The accumulated NSRs entering and leaving block \(l\) are
\begin{equation}
    R_l=\frac{\|H_l^{\mathrm{q}}-H_l^{\mathrm{fp}}\|_F^2}{\|H_l^{\mathrm{fp}}\|_F^2},
    \qquad
    R_{l+1}=\frac{\|T_l\|_F^2}{\|F_l^{\mathrm{fp}}(H_l^{\mathrm{fp}})\|_F^2}.
\end{equation}
Using the same output energy of full-precision models, the propagated, source and interaction components are
\begin{equation}
    R_l^{\mathrm{prop}}=\frac{\|P_l\|_F^2}{\|F_l^{\mathrm{fp}}(H_l^{\mathrm{fp}})\|_F^2},
    \quad
    Q_l=\frac{\|C_l\|_F^2}{\|F_l^{\mathrm{fp}}(H_l^{\mathrm{fp}})\|_F^2},
    \quad
    R_l^I=\frac{\|I_l\|_F^2}{\|F_l^{\mathrm{fp}}(H_l^{\mathrm{fp}})\|_F^2}.
\end{equation}
For \(R_l>0\), the propagation gain is
\begin{equation}
    G_l=\frac{R_l^{\mathrm{prop}}}{R_l}.
\end{equation}
Thus, \(G_l>1\) indicates amplification of the accumulated error, \(G_l<1\) indicates attenuation, and \(G_l\approx1\) indicates that its relative energy is largely preserved through the block.

\paragraph{Layer-wise recurrence.}
Expanding the squared norm of the exact decomposition gives
\begin{equation}
    \|T_l\|_F^2=\|P_l\|_F^2+\|C_l\|_F^2+\|I_l\|_F^2+2\langle P_l,C_l\rangle+2\langle P_l,I_l\rangle+2\langle C_l,I_l\rangle .
\end{equation}
After division by the output energy of full-precision models, the exact layer-wise relation is
\begin{equation}
    R_{l+1}=R_l^{\mathrm{prop}}+Q_l+D_l=G_lR_l+Q_l+D_l,
\end{equation}
where
\begin{equation}
    D_l=\frac{\|I_l\|_F^2+2\langle P_l,C_l\rangle+2\langle P_l,I_l\rangle+2\langle C_l,I_l\rangle}{\|F_l^{\mathrm{fp}}(H_l^{\mathrm{fp}})\|_F^2}.
\end{equation}
The term \(D_l\) contains the interaction energy and the signed overlaps among \(P_l\), \(C_l\) and \(I_l\). It can therefore be negative and is not suitable for reporting a non-negative relative energy share.

To test how much of the accumulated NSR is explained by propagation and source error alone, we omit \(D_l\). Because the full-precision and quantized models receive identical initial inputs, \(R_0=0\) and the first block gives \(\widehat{R}_1=Q_0\). The remaining blocks are evaluated recursively as
\begin{equation}
    \widehat{R}_{l+1}=G_l\widehat{R}_l+Q_l,
    \qquad l=1,\ldots,L-1.
\end{equation}
For \(L\) transformer blocks indexed by \(l=0,\ldots,L-1\), the reconstructed NSR at the output of the final block is
\begin{equation}
    \widehat{R}_L=Q_{L-1}+\sum_{l=0}^{L-2}\left(\prod_{j=l+1}^{L-1}G_j\right)Q_l.
\end{equation}
This form makes explicit that the source NSR from the final block contributes directly, whereas source errors introduced earlier are scaled by the propagation gains of all subsequent blocks.

The notation used in the decomposition and recurrence is summarized in Supplementary Table~\ref{tab:app_source_propagation_quantities}.

\begin{table}[!t]
    \centering
    \caption{
        \textbf{Notation for the source--propagation analysis.}
    }
    \label{tab:app_source_propagation_quantities}
    \begin{tabularx}{\textwidth}{@{}llX@{}}
        \toprule
        Symbol & Term & Interpretation \\
        \midrule
        \(R_l\) & Accumulated NSR entering block \(l\) & Error present before the block \\
        \(R_{l+1}\) & Observed accumulated NSR after block \(l\) & Total output error after the block \\
        \(R_l^{\mathrm{prop}}\) & Propagated NSR & Accumulated error after passing through the full-precision block \\
        \(Q_l\) & Source NSR & Error introduced by quantizing the current block \\
        \(R_l^I\) & Interaction NSR & Non-negative energy of the interaction term \\
        \(G_l\) & Propagation gain & Amplification or attenuation of accumulated error \\
        \(D_l\) & Residual recurrence term & Interaction energy and signed overlaps omitted from the reduced recurrence \\
        \(\widehat{R}_l\) & Reconstructed accumulated NSR & Accumulated NSR reconstructed from measured \(G_l\) and \(Q_l\) \\
        \bottomrule
    \end{tabularx}
\end{table}

\paragraph{Aggregation.}
The global quantities in panels a--c of the main-text Fig.~4 and Supplementary Figs.~\ref{fig:app_source_propagation_hellaswag} and~\ref{fig:app_source_propagation_gsm8k} are Frobenius-energy ratios formed after summing squared values over all valid tokens and representation dimensions. For panel a, the relative shares of \(R_l^{\mathrm{prop}}\), \(Q_l\) and \(R_l^I\) are computed at each layer, averaged within the early, middle and late thirds of normalized depth, and re-normalized to sum to one. We report \(R_l^I\), rather than \(D_l\), because \(R_l^I\) is a non-negative energy ratio. Panels d and e report the channel median of accumulated NSR and source NSR at each layer.

\subsection{Source--propagation analysis across tasks}
\label{app:source_propagation_across_tasks}

Main-text Fig.~4 reports the analysis on WikiText-2. To examine whether the observed accumulation behavior depend on the task, we repeat the analysis on HellaSwag and GSM8K while retaining the same Qwen3 models, GPTQ-W3 quantization and aggregation procedure. Data preparation for the three datasets is described in Supplementary Table~\ref{tab:app_internal_inputs}.

\begin{figure}[p]
    \centering
    \includegraphics[width=\textwidth]{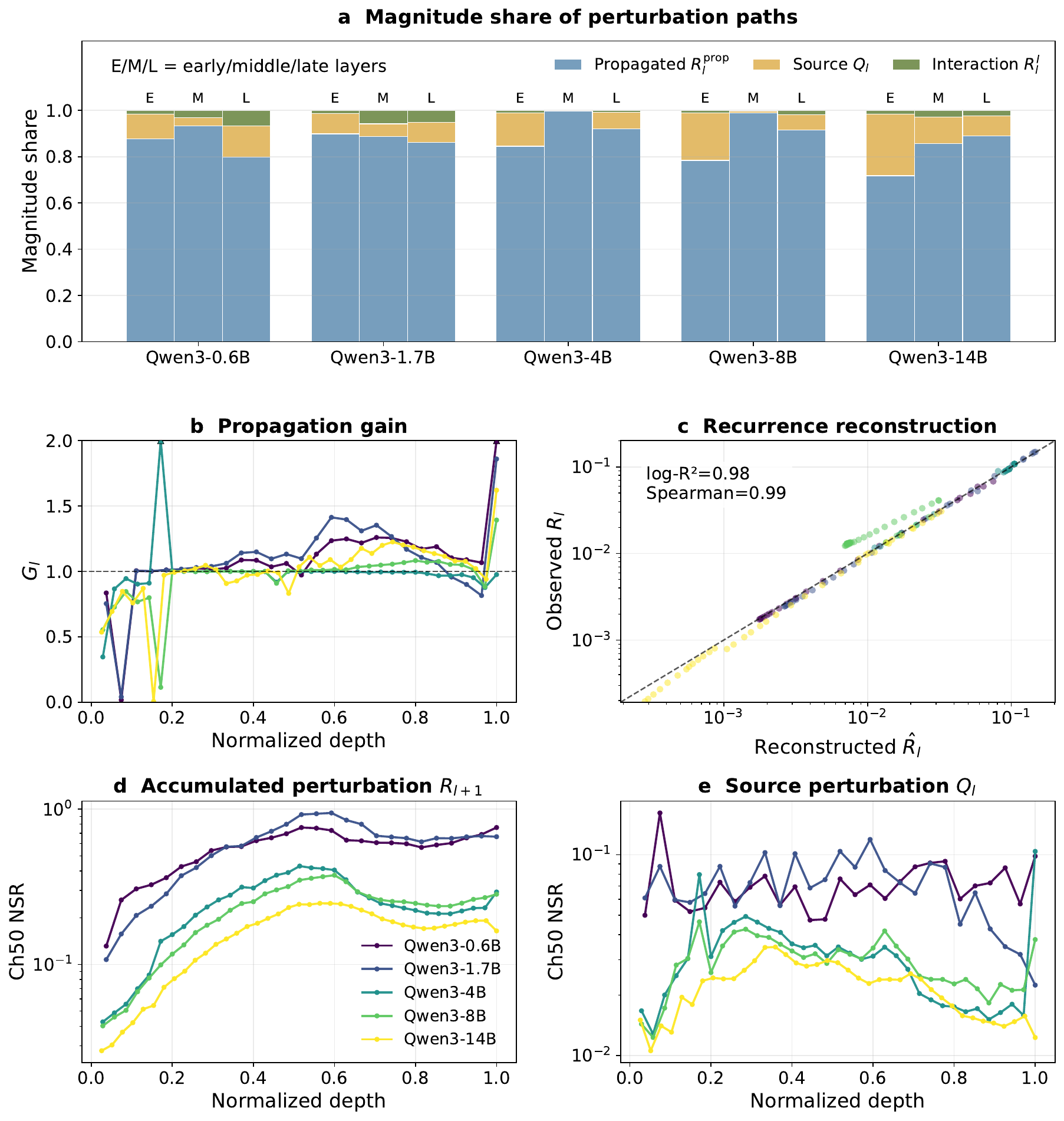}
    \caption{
        \textbf{Cross-layer source--propagation analysis on HellaSwag.}
        Qwen3 models from 0.6B to 14B are quantized with GPTQ-W3.
        \textbf{a}, Relative shares of propagated NSR \(R_l^{\mathrm{prop}}\), source NSR \(Q_l\) and interaction NSR \(R_l^I\) within the early, middle and late thirds of normalized depth.
        \textbf{b}, Propagation gain \(G_l\) across depth; the dashed line marks \(G_l=1\), and values outside the displayed range are clipped.
        \textbf{c}, Reconstructed accumulated NSR \(\widehat{R}_l\) versus observed accumulated NSR \(R_l\).
        \textbf{d}, Channel-median accumulated NSR \(R_{l+1}\).
        \textbf{e}, Channel-median source NSR \(Q_l\).
        Panels \textbf{a}--\textbf{c} use global Frobenius-energy ratios, whereas panels \textbf{d} and \textbf{e} report the median across channels at each layer.
    }
    \label{fig:app_source_propagation_hellaswag}
\end{figure}

\begin{figure}[p]
    \centering
    \includegraphics[width=\textwidth]{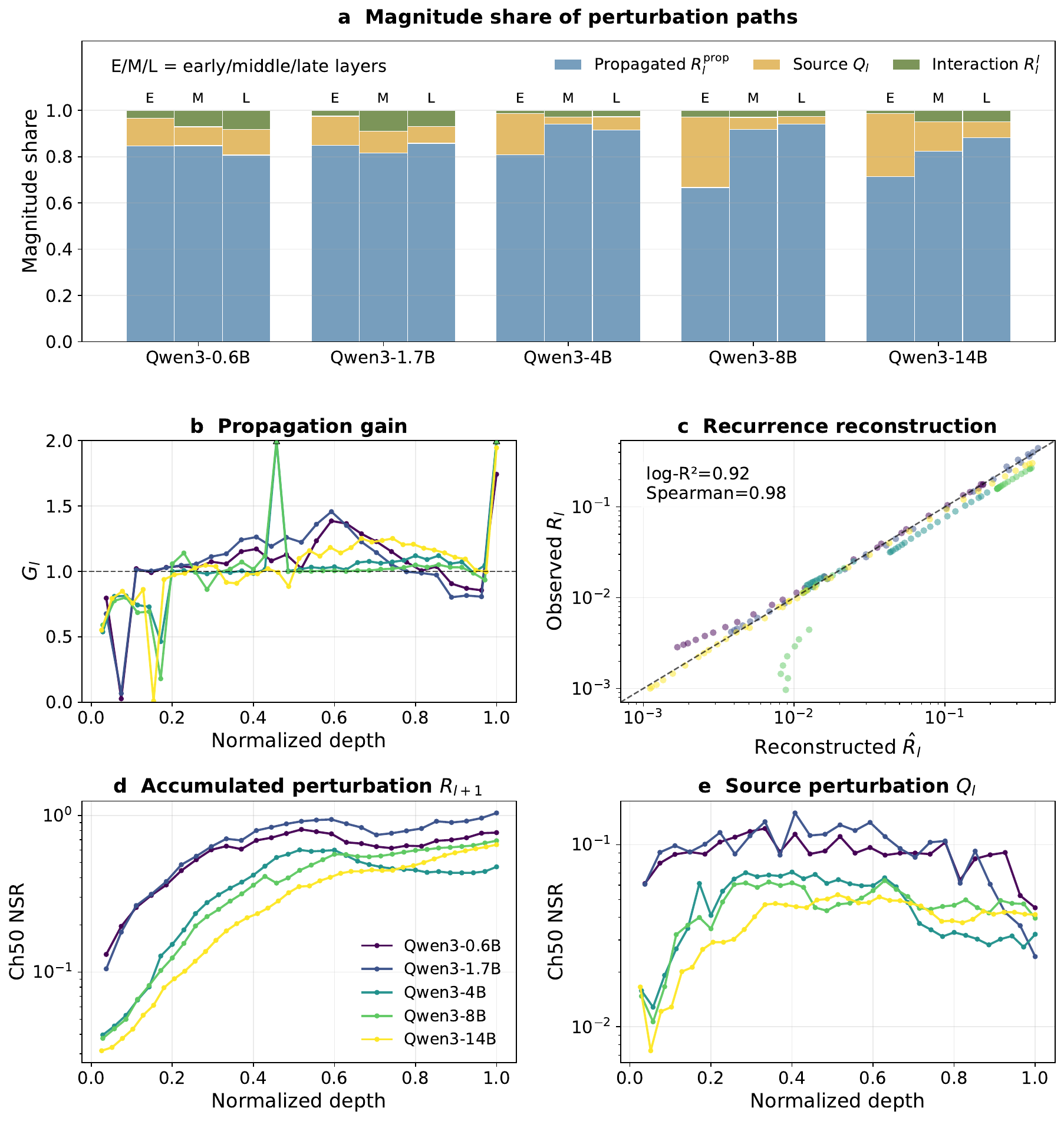}
    \caption{
        \textbf{Cross-layer source--propagation analysis on GSM8K.}
        Qwen3 models from 0.6B to 14B are quantized with GPTQ-W3.
        \textbf{a}, Relative shares of propagated NSR \(R_l^{\mathrm{prop}}\), source NSR \(Q_l\) and interaction NSR \(R_l^I\) within the early, middle and late thirds of normalized depth.
        \textbf{b}, Propagation gain \(G_l\) across depth; the dashed line marks \(G_l=1\), and values outside the displayed range are clipped.
        \textbf{c}, Reconstructed accumulated NSR \(\widehat{R}_l\) versus observed accumulated NSR \(R_l\).
        \textbf{d}, Channel-median accumulated NSR \(R_{l+1}\).
        \textbf{e}, Channel-median source NSR \(Q_l\).
        Panels \textbf{a}--\textbf{c} use global Frobenius-energy ratios, whereas panels \textbf{d} and \textbf{e} report the median across channels at each layer.
    }
    \label{fig:app_source_propagation_gsm8k}
\end{figure}

Supplementary Figs.~\ref{fig:app_source_propagation_hellaswag} and~\ref{fig:app_source_propagation_gsm8k} capture the main observations. Propagated error accounts for the largest share of the accumulated error across model scales and depth stages, while the interaction energy remains comparatively small. In the middle layers, Qwen3-0.6B and Qwen3-1.7B exhibit more sustained amplification than the larger models, whose propagation gains remain closer to one. The channel-wise curves likewise show lower source and accumulated NSR for the larger models across most layers, although adjacent model scales are not strictly ordered.

The reduced recurrence also captures the overall accumulation pattern on both tasks. For HellaSwag, the reconstructed and observed accumulated NSR achieve a log-scale \(R^2\) of 0.98 and a Spearman correlation of 0.99; for GSM8K, the corresponding values are 0.92 and 0.98. The deviations at some layers, particularly for Qwen3-8B on GSM8K, show that the omitted term \(D_l\) can remain relevant locally even when the propagation and source terms explain the dominant trend.

\clearpage

